\documentclass[twoside]{article}

\usepackage[accepted]{aistats2025halarxiv}

\usepackage[round]{natbib}

\usepackage{titlesec}
\usepackage{multirow}
\usepackage{bm}
\usepackage{enumitem}
\usepackage{hyperref}
\usepackage{cleveref}
\usepackage{xcolor}
\usepackage{graphicx}
\usepackage[outdir=./]{epstopdf}
\usepackage{caption}
\usepackage{subcaption}
\usepackage{tabularx}
\usepackage{amsfonts}
\usepackage{amsmath}
\usepackage{amssymb}
\usepackage{mathabx}
\usepackage{stfloats}
\usepackage{titlesec}
\usepackage{amsthm}
\usepackage{bbold}
\usepackage{booktabs}
\usepackage{algorithm}
\usepackage{algpseudocode}

\newtheorem{definition}{Definition}

\DeclareMathOperator*{\argmin}{argmin}
\newcommand{\RRMSE}{\mathrm{RRMSE}}

\begin{document}



\twocolumn[

\aistatstitle{Learning signals defined on graphs with optimal transport and Gaussian process regression}

\aistatsauthor{Raphaël Carpintero Perez\textsuperscript{1,2} \: \: \: Sébastien Da Veiga\textsuperscript{3} \: \: \: Josselin Garnier\textsuperscript{2} \: \: \: Brian Staber \textsuperscript{1}}
\aistatsaddress{ \textsuperscript{1} Safran Tech, Digital Sciences \& Technologies, 78114 Magny-Les-Hameaux, France\\ \textsuperscript{2} Centre de Mathématiques Appliquées, Ecole polytechnique, Institut Polytechnique de Paris, 91120 Palaiseau, France\\
\textsuperscript{3} Univ Rennes, Ensai, CNRS, CREST - UMR 9194, F-35000 Rennes, France}
]

\begin{abstract}
In computational physics, machine learning has now emerged as a powerful complementary tool to explore efficiently candidate designs in engineering studies. Outputs in such supervised problems are signals defined on meshes, and a natural question is the extension of general scalar output regression models to such complex outputs. Changes between input geometries in terms of both size and adjacency structure in particular make this transition non-trivial. In this work, we propose an innovative strategy for Gaussian process regression where inputs are large and sparse graphs with continuous node attributes and outputs are signals defined on the nodes of the associated inputs. The methodology relies on the combination of regularized optimal transport, dimension reduction techniques, and the use of Gaussian processes indexed by graphs. In addition to enabling signal prediction, the main point of our proposal is to come with confidence intervals on node values, which is crucial for uncertainty quantification and active learning. 
Numerical experiments highlight the efficiency of the method to solve real problems in fluid dynamics and solid mechanics.
\end{abstract}

\section{INTRODUCTION}
Many problems in computational physics rely on solving partial differential equations on a domain with given geometry using the finite element method (FEM). Although much appreciated, FEM often involves meshes with highly refined discretization, which quickly becomes computationally intensive even with parallel computing. In particular, exploring how changes in the geometry impact some key quantities of interest computed by FEM is an everyday task in engineering for design studies. Due to the associated computational cost, machine learning (ML) is a natural candidate to accelerate such design exploration: starting from an initial database of FEM simulations, a supervised model is trained to predict the FEM outputs from its inputs and is ultimately used as a proxy to evaluate new geometries with a negligible cost. But in this context, the supervised learning task actually involves inputs given as meshes, which can be modeled as graphs with continuous node attributes, different numbers of nodes and edges. In addition, the outputs can be scalar values but also physical quantities of interest defined on each node of the input graph, which we refer to as \emph{signals defined on graphs} or \emph{fields}. 

These two specificities give rise to a supervised learning problem which is not in a standard form and thus calls for dedicated algorithms. A prominent example are graph neural networks (GNNs), which have shown remarkable success for this type of application by iteratively aggregating and transforming information from node neighbors. However they still have inherent limitations when the dataset is small and when it comes to predicting output values with associated uncertainties. ML models offering predictive uncertainties play a key role in a wide range of critical industrial applications as they can certify the quality of results \citep{conformal_jaber}, can assist sequential design of experiments \citep{computer_exp1, computer_exp2} or can be plugged into Bayesian optimization workflows \citep{ego}. Gaussian process (GP) regression is quite a popular approach in this small data setting with required uncertainties, but unfortunately suffers from shortcomings for FEM applications: even if the use of graph kernels can handle inputs defined as graphs, GPs do not generalize easily to complex outputs such as signals on graphs with varying and large number of nodes.

\begin{figure*}[ht]
\begin{center}
\includegraphics[width=1.\textwidth]{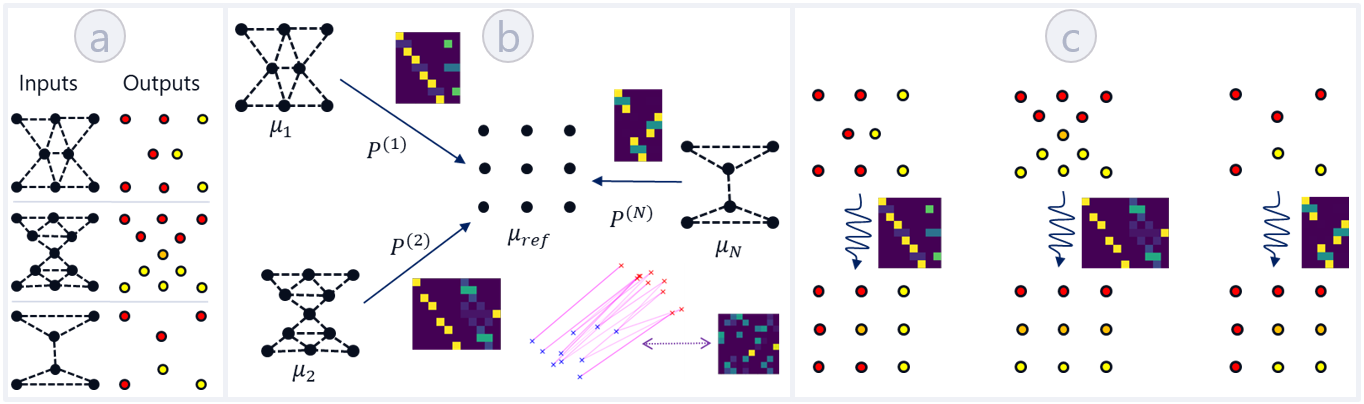}
\end{center}
\caption{Summary of our approach. a) Inputs = Graphs. Outputs = Fields defined on the nodes. b) Step 1: obtaining transport plans to a reference measure. c) Step 2: Transferring signals to the reference measure.}
\label{fig:swwl}
\end{figure*}

In this article, we introduce the Transported Output Signal GP regression (TOS-GP) that makes it possible to extend scalar Gaussian process regression to output signals thanks to a regressor-agnostic transformation of the outputs combined with dimension reduction. In a preliminary step, each input graph is pre-processed and encoded as an empirical measure supported on the nodes to which we assign their continuous attributes and those of their neighbors.  We then select a reference measure and compute optimal transport plans between all input empirical measures and this reference: this step is crucial since it yields a transformation allowing to transport any information from any graph with any number of nodes to a common space, the reference measure support. In particular, each output signal is transported with the plan corresponding to its input graph:  intuitively, this means that if nodes on two graphs match, then the signals defined on them should match too, see Figure \ref{fig:swwl} for an illustration. The supervised learning problem now writes in an amenable form, although possibly with a large number of new outputs equal to the size of the reference measure. The last step thus consists in applying a dimension reduction technique to finally learn a few independent scalar-valued GPs, taking graphs as inputs, for which we use the Sliced Wasserstein Weisfeiler-Lehman (SWWL) graph kernel \cite{swwl}. 

The article is organized as follows. Section \ref{sec:preliminaries} first introduces the problem setting and recalls GP regression. Related work are discussed in Section \ref{section:related}. The TOS-GP methodology is described in Section \ref{section:optimally_transported_signals}. Numerical experiments are then performed in Section \ref{section:exp} and Section \ref{section:varying_topologies} with a focus on real datasets coming from FEM simulations.

\section{PRELIMINARIES} \label{sec:preliminaries}

We consider the task of learning a function $f : \mathcal{X} \rightarrow \mathcal{Y}$. $\mathcal{X}$ denotes a space of undirected graphs with continuous node attributes. Each element in $\mathcal{X}$ can be written as $X=(V,E,w,\mathbf{F})$ where $V$ is the set of $|V|$ nodes, $E$ is a set of paired nodes, whose elements are called edges, and the function $w : E \rightarrow \mathbb{R}$ assigns the edge weights. The $d$-dimensional attributes of the nodes are gathered in the $|V| \times d$ matrix $\mathbf{F} = (\mathbf{F}_u)_{u \in V}$. The output space $\mathcal{Y}$ is a set of signals defined on graphs of $\mathcal{X}$, that is to say:\\ $ \mathcal{Y} = \bigcup\limits_{ \substack{X=(V,E,w,\mathbf{F})\in \mathcal{X}}} \{ Y: V \rightarrow \mathbb{R} \}$. By abuse of notation, the signal can be denoted as the vector $\mathbf{Y}=(Y(1), \cdots, Y(|V|))^T$. Remark that the outputs are permutation equivariant: for each permutation of the input nodes $(u_1,\cdots, u_{|V|})$, the output signal needs to be permuted equivalently.

We assume that we are given a dataset $\mathcal{D}$ consisting of $N$ observations $\mathcal{D} = \{ (X^{(i)}, \mathbf{Y}^{(i)}) \}^{N}_{i=1}$, where the inputs may differ in terms of numbers of nodes and adjacency matrices, \textit{i.e.}, we may have $|V^{(i)}|\neq |V^{(j)}|$ and $E^{(i)} \neq E^{(j)}$ for some $(i, j) \in \{1, \ldots, N\}^2$.

For this problem, we focus on GP regression, which is a popular Bayesian approach for supervised learning in engineering \citep{gpml,gramacy2020surrogates}. In the particular case of a single output, we try to learn $g: \mathcal{X} \rightarrow \mathbb{R}$ from noisy training observations  $y^{(i)} = g(X^{(i)}) + \epsilon^{(i)}$ for $i \in \{1, \ldots, N\}$ of $g:{\mathcal X}\to \mathbb{R}$ at input locations $\mathbf{X} = (X^{(i)})_{i=1}^N$, where $\epsilon^{(i)} \sim \mathcal{N}(0,\eta^2)$ is an additive i.i.d. Gaussian noise. Let $\mathbf{g_*}:=(g({X}_{*}^{(i)}))_{i=1}^{N^*}$ be the values of $g$ at new test locations $\mathbf{X}_{*}=({X}_{*}^{(i)})_{i=1}^{N^*}$. 

A zero-mean GP prior is placed on the function $g$ (adding a constant mean is immediate, and has no impact in practice for our use cases). It follows that the joint distribution of the observed target values and the function values at the test locations writes (see, \textit{e.g.}, \citet{gpml})
\begin{equation}
    \begin{bmatrix}
    \mathbf{y}\\
    \mathbf{g}_*
    \end{bmatrix}
    \sim \mathcal{N} 
    \left( 
    \mathbf{0}, 
    \begin{bmatrix}
    \mathbf{K} + \eta^2 \mathbf{I} & \mathbf{K}_*^T\\
    \mathbf{K}_* & \mathbf{K}_{**}
    \end{bmatrix}
    \right)\,,
\end{equation}

where $\mathbf{K}$, $\mathbf{K}_{**}$, $\mathbf{K}_{*}$ are the train, test and test/train Gram matrices, respectively.
The posterior distribution of $\mathbf{g}_*$, obtained by conditioning the joint distribution on the observed data, is also Gaussian: $\mathbf{g}_* | \mathbf{X}, \mathbf{y}, \mathbf{X}_{*} \sim \mathcal{N}(\mathbf{\bar{m}},\mathbf{\bar{\Sigma}})$ with mean and covariance given by
\begin{align}
\mathbf{\bar{m}}&=\mathbf{K}_*(\mathbf{K}+\eta^2\mathbf{I})^{-1} \mathbf{y}\,, \\
\mathbf{\bar{\Sigma}}&=\mathbf{K}_{**}-\mathbf{K}_* (\mathbf{K}+\eta^2 \mathbf{I})^{-1}\mathbf{K}_*^T\,.
\end{align}
The mean of the posterior distribution is used as a predictor, and predictive uncertainties can be obtained through the covariance matrix. GP regression requires a positive definite kernel function $k : \mathcal{X}\times\mathcal{X}\rightarrow \mathbb{R}$. When $\mathcal{X}$ is a space of graphs, we rely on graph kernels. 

When the output space $\mathcal{Y}$ consists of signals on graphs, the learning problem possesses several characteristics that require special attention:
\begin{itemize}[topsep=0pt]
\itemsep0em 
    \item Inputs $X^{(1)}, \cdots, X^{(N)}$ can have different sizes, so the outputs $\mathbf{Y}^{(1)}, \cdots, \mathbf{Y}^{(N)}$ do not have a fixed size common to all samples
     \item There is no natural ordering of the output dimensions, leading to the permutation equivariance of the signals
     \item The output dimension can be very large 
\end{itemize}

\section{RELATED WORK} \label{section:related}

\paragraph{GNNs and morphing.}
GNNs are indisputably the reference to build predictions on all the nodes of input graphs by relying on the message passing framework introduced by \cite{gnn_message_passing} and extended by \cite{gnn_message_passing2}, with recent advances targeting specifically solutions to physical systems, either by incorporating physical knowledge \citep{gnn_fno} or by designing efficient GNNs architectures \citep{gnn_mgn_pfaff2020}. GNNs are still limited by the need of a large sample size, and require a specific treatment to get access to uncertainties as elaborated in \citep{surveyUQNN}. Alternatively, the Mesh Morphing GP approach proposed by \cite{mmgp} consists in morphing the meshes to a reference shape, on which the output fields are then interpolated, followed by dimension reduction. This is close in sprit to our proposal, but this method primarily relies on morphing, which is limited to mesh data only, and meshes with the same topology. On the contrary our method is completely generic since we leverage optimal transport instead of morphing.

\paragraph{From single to multi-output prediction.}
When outputs are vectors in $\mathbb{R}^d$, it is possible to consider multi-output GPs \citep{kernels_for_vv_functions} based on vector-valued RKHS \citep{vv_rkhs}, with the famous Linear Model of Coregionalization (LMC) of \cite{lmc}, Intrinsic Coregionalization Model (ICM) of \cite{icm} and Convolved Gaussian Processes of \cite{alvarez2011computationally}. However these approaches are not really suitable when the output dimension is large. When the output actually is the discretization of a continuous phenomenon, corresponding to infinite dimensional outputs, operator-valued kernels \citep{operator_valued_kernels_kadri, operator_valued_gp_kadri} are a natural extension of vector-valued kernels when the output space writes as a space of functions. Instead, one can also use dimension reduction in the output space to decompose the problem into independent sub-problems while managing the uncertainty quantification \citep{dim_reduction_survey}. Recent applications show the use of Singular Value Decomposition \citep{dim_reduction_svd}, wavelet transforms \citep{dim_reduction_wavelets} and autoencoders \citep{dim_reduction_autoencoders}. Unfortunately, all methods mentioned so far assume that the outputs are defined on the same domain across the samples. An exception exists, when it is possible to measure a similarity between structured outputs via a positive definite kernel: the output kernel regression framework \citep{kde, jkm, structured_output_learning_kadri, iokr} performs prediction in the associated output RKHS instead. But ultimately it requires to solve a non-trivial pre-image problem at test time, requiring a minimization over the output space which is not tractable in the applications we consider. 
%which is especially cumbersome when dealing with graphs.

\paragraph{Graph signal processing.}
We can also think of dimension reduction techniques based on the graph Laplacian like graph signal processing \citep{the_emerging_fields_of_gsp}, with many recent applications in ML \citep{gsp_for_ml_review, fourier_could_be_a_data_scientist}. Such approaches usually learn functions defined on the nodes of a common graph (an input is a node, and an output is a scalar) either by regularizing with the Laplacian matrix \citep{gp_on_graphs} or by using spectral techniques \citep{gp_on_graphs_via_spectral}. The use of Laplacian eigenvectors however suffers from indeterminacies due to the choices of signs and basis, as described by \cite{sign_and_basis_invariant_gnn}. In particular, comparing representations of signals defined on graphs of different sizes by projecting them onto their respective eigenvectors is irrelevant since comparing two eigenvectors of different sizes makes little sense. %But none of them is able to handle graphs with different Laplacian matrices, except for neural networks.

 \paragraph{Optimal transport.}
Using optimal transport to handle different domain distributions \citep{computational_ot, montesuma2023recent} is not at all new, for example color or texture transfer that involves transport plans between histograms \citep{ferradans2014regularized}. Optimal transport has also seen some applications in the domain of computational physics to interpolate between fields thanks to optimal transport between Gaussian models \citep{iollo2022mapping} or to register point clouds \citep{shen2021accurate}. In the context of domain adaptation \citep{ot_for_domain_adaptation}, the recent approach of \cite{jdot} can handle changes not only in the input train and test distributions, but also in the output train and test domains by learning simultaneously a predictive function and the optimal transport plan between joint distributions. Even if they share similar ideas, these approaches cannot be used directly in our case since the domains are different for each input, and the output signals to be predicted cannot be used to obtain transport plans. To our knowledge, we propose the first approach to reuse a transport plan defined in feature space in order to change the support of output fields.

\section{OPTIMALLY TRANSPORTED SIGNALS}
\label{section:optimally_transported_signals}
Our TOS-GP methodology relies on the entropy-regularized Wasserstein distance to transfer the signals to a reference measure. Given this reference measure, we find the regularized optimal transport plans between input measures and the reference measure. We then transfer the signals to be expressed on the support of the reference measure using the latter transport plans. The transferred signals now have a fixed order and a fixed size, on which dimension reduction techniques can now be applied. We then learn the low-dimensional embeddings using independent GPs. Algorithm \ref{algo:tos-gp-train} and algorithm \ref{algo:tos-gp-test} respectively summarize the training and test stages.

The preliminary step consists in encoding the inputs of our original problem so that we can use optimal transport. Recall that $X^{(1)}, \cdots, X^{(N)}$ are $N$ graphs 
%with respective feature vectors $\mathbf{F}^{(1)}, \cdots, \mathbf{F}^{(N)}$ of sizes $n_1, \cdots, n_N$
with $n_1, \cdots, n_N$ nodes and with respective feature matrices $\mathbf{F}^{(1)}, \cdots, \mathbf{F}^{(N)}$. 
In order to leverage the adjacency structure of the graphs, we incorporate the continuous Weisfeiler-Lehman (WL) embeddings \citep{wwl, swwl} to 
%the feature matrix $\mathbf{F} = (\mathbf{F}_u)_{u \in V}$ of each node, 
the feature vector $\mathbf{F}_u^{(i)}$ of each node $u$, 
whose dimension becomes $d\times (H+1)$ where $H$ is the number of continuous WL iterations. Each input graph $X^{(i)}$ is finally encoded as an input empirical measure $\mu^{(i)} = \frac{1}{n_i} \sum_{u=1}^{n_i} \delta_{\mathbf{F}^{(i)}_u}$ supported by its features. 

\paragraph{Regularized Wasserstein distance.}

We first recall the definition of the Wasserstein distance for arbitrary measures on $\mathbb{R}^s, s\geq 1$ (in our setting, $s=d\times (H+1)$).

\begin{definition}[Wasserstein distance]
\label{def:wasserstein}
Let $s \geq 1$ be an integer, $r \geq 1$ be a real number, and $\mu, \nu$ be two probability measures on $\mathbb{R}^s$ having finite moments of order $r$. The $r$-Wasserstein distance is defined as
\begin{equation}
W_r(\mu, \nu) := \bigg( \inf\limits_{\pi \in \Pi(\mu, \nu)} \int\limits_{\mathbb{R}^s\times\mathbb{R}^s} \|\mathbf{x}-\mathbf{y}\|^r d\pi(\mathbf{x},\mathbf{y}) \bigg)^{\frac{1}{r}}\,,  
\end{equation}
where $\Pi(\mu, \nu)$ is the set of all probability measures on $\mathbb{R}^s\times\mathbb{R}^s$ whose marginals w.r.t.
the first and second variables are respectively $\mu$ and $\nu$ and $\|\cdot\|$ stands for the Euclidean norm on $\mathbb{R}^s$.
\end{definition}

In the following, we consider only discrete measures $\mu = \frac{1}{n} \sum_{u=1}^{n} \delta_{\mathbf{x}_u}$ and $\nu = \frac{1}{m} \sum_{v=1}^{m} \delta_{\mathbf{x}'_v}$ supported respectively on the points $\mathbf{x}_1, \cdots, \mathbf{x}_n$ and $\mathbf{x}'_1, \cdots, \mathbf{x}'_m \in \mathbb{R}^s$ (by abuse of language, we sometimes refer to the size of a measure as the size of its support). We denote by  $U(n,m) = \{ \mathbf{P} \in \mathbb{R}^{n\times m}_+ : \mathbf{P}\mathbb{1}_m = \frac{1}{n}\mathbb{1}_n \text{ and } \mathbf{P}^T \mathbb{1}_n = \frac{1}{m}\mathbb{1}_m \}$ the set of admissible coupling matrices with mass preservation, where $\mathbb{1}_n$ is the column vector composed of ones. We denote by $\mathbf{C}^{\mu, \nu} := (c(\mathbf{x}_u, \mathbf{x}'_v))_{u,v}$ the $n \times m$ cost matrix, where $c:\mathbb{R}^s\times \mathbb{R}^s \rightarrow \mathbb{R}$ is the cost function ($c(\mathbf{x},\mathbf{x}') = \|\mathbf{x}-\mathbf{x}'\|^2$ for the $r$-Wasserstein distance with $r=2$).
% d change en s
% i change en u

The Kantorovich formulation of optimal transport writes as follows:
\begin{equation}
\label{eq:Kantorovitch}
L(\mu, \nu) := \min_{\mathbf{P}\in U(n,m)} \langle \mathbf{C}^{\mu,\nu}, \mathbf{P} \rangle  .
\end{equation}
When $n=m$, then there exists an optimal solution for (\ref{eq:Kantorovitch}) which is a scaled permutation matrix.
To obtain approximate solutions to the original transport problem (\ref{eq:Kantorovitch}), one can add an entropic regularization term: 
\begin{align}
\begin{split}
\label{eq:reg-OT}
\mathcal{L}_\lambda(\mu, \nu, \mathbf{P}) &:= \langle \mathbf{C}^{\mu,\nu}, \mathbf{P}\rangle - \lambda H(\mathbf{P}) , \\
L_{\lambda}(\mu, \nu) &:=
\min_{\mathbf{P} \in U(n,m)} \mathcal{L}_\lambda(\mu, \nu, \mathbf{P}) ,
\end{split}
\end{align}
where the discrete entropy of a coupling matrix is defined as $H(\mathbf{P}) := -\sum_{u,v} P_{u,v} (\log(P_{u,v})-1)$, and $\lambda \geq 0$ is the regularization parameter. When $\lambda > 0$, the objective is a $\lambda$-strongly convex function that has a unique minimizer $P_\lambda := \argmin_{\mathbf{P} \in U(n,m)} \mathcal{L}_\lambda(\mu, \nu, \mathbf{P})$. Solving the regularized problem has a two-fold benefit in our case. First, the $O(n^3 \log(n))$ complexity of the evaluation of the Wasserstein distance can be reduced to $O(n^2 \log(n))$ with entropic regularization, and the computation can be accelerated using GPU devices thanks to Sinkhorn iterations (the higher the coefficient, the quicker it will run). On the other hand, obtaining a more diffuse transport plan can be beneficial for our signal transfer application as it can induce a smoothing depending on the regularization parameter $\lambda$.

\paragraph{Signal transfer with optimal transport plans.}
\label{section:field_transfer_with_optimal_transport_plans}
Given a reference measure $\mu_{\textrm{ref}}$ supported on $n_{\textrm{ref}}$ points, and a common regularization parameter $\lambda$, we compute the entropy-regularized transport plan $\mathbf{P}_\lambda^{(i)}$ between each input measure $\mu^{(i)}$ and $\mu_{\textrm{ref}}$:
\begin{equation}
\label{eq:transport_plans}
\mathbf{P}^{(i)} = \argmin_{\mathbf{P} \in U(n_i,n_{\textrm{ref}})} \mathcal{L}_\lambda(\mu^{(i)}, \mu_{\textrm{ref}}, \mathbf{P}).
\end{equation}
Such transport plans provide matches between several points of the involved measures, and the proportion of the masses that need to be split between them. A key ingredient of our method is to use the same transport plans in the output space, in order to obtain a representation of the signals with a fixed support given by the reference measure, called the \emph{OT-transferred fields} given by:
\begin{equation}
\label{eq:OT-transferred-signals}
\mathbf{T}^{(i)} = (n_{\textrm{ref}} \mathbf{P}^{(i)})^T \mathbf{Y}^{(i)} \in \mathbb{R}^{n_{\textrm{ref}}}.
\end{equation}
Crucially, we can transfer back signals defined on the reference measure to their original support by applying the transposed of the transfer operator:
\begin{equation}
\label{eq:OT-reconstructed-signals}
\tilde{\mathbf{Y}}^{(i)} = (n_i \mathbf{P}^{(i)}) \mathbf{T}^{(i)} \in \mathbb{R}^{n_i}.
\end{equation}
Remark that this reconstruction cannot be exact as soon as $n_i \neq n_{\textrm{ref}}$ or $\lambda>0$.

\paragraph{Dimension reduction.}
Since the transferred signals are now expressed on the same reference measure, any dimension reduction technique $\mathcal{M}_{DR}$ can be applied to $\{ \mathbf{T}^{(i)} \}^{N}_{i=1}$ in order to obtain some low-dimensional embeddings $\{ \mathbf{C}^{(i)} \}^{N}_{i=1} = \mathcal{M}_{DR}(\{\mathbf{T}^{(i)}\}_{i=1}^N)$, where $\mathbf{C}^{(i)} = (C^{(i)}_j)_{j=1}^Q  \in \mathbb{R}^{Q}$. For instance, using Principal Component Analysis (PCA) \citep{pca}, $\mathbf{C}^{(i)}$ are the first $Q$ PCA coefficients of each transferred signal obtained by multiplication with the projection matrix $\mathbf{E}^T$ containing the first $Q$ PCA eigenvectors.

\paragraph{Single-output GP regression.}
Approximating $f: \mathcal{X} \rightarrow \mathcal{Y}$ can now be broken down to learning $Q$ independent functions: $g^{(j)}: \mathcal{X} \rightarrow \mathbb{R}$ using the respective datasets $\mathcal{D}_j = \{ (X^{i}, C^{(i)}_j) \}^{N}_{i=1}$ for $j=1,\cdots,Q$. 
Any positive definite graph kernel \citep{a_survey_on_gk, gk_a_survey} can be used, but here we choose to use the SWWL kernel of \cite{swwl}, which has shown good performance for scalar prediction when inputs are graphs with continuous attributes.

Alternatively, multi-output Gaussian processes like the ICM \citep{icm} could be use in place of Gaussian process regression to explicitely incorporate the correlation matrix between PCA coefficients, but it would be limited to very low values of $Q$.
%However, even with a prior dimension reduction, the covariance matrix would add $\frac{Q(Q+1)}{2}$ parameters to estimate, which turns out to be problematic even for small values of $Q$.}

\begin{algorithm}[H]
  \caption{TOS-GP, training phase}\label{algo:tos-gp-train}
  \begin{algorithmic}[1]
  \Require{Train dataset $\mathcal{D} = \{ (X^{(i)}, \mathbf{Y}^{(i)}) \}^{N}_{i=1}$, reference measure $\mu_{\textrm{ref}}$, regularization parameter $\lambda$} 
  \Ensure{(Linear) dimension reduction model $\mathcal{M}_{DR}$, list of $Q$ GP models $\{ \mathcal{M}_{gp,j}\}_{j=1}^Q$}
    %\Procedure{Euclid}{$a,b$}\Comment{The g.c.d. of a and b}
  \For{$i=1, \cdots, N$}
  \State $\mu^{(i)} \gets \frac{1}{n_i} \sum_{u=1}^{n_i} \delta_{\mathbf{F}^{(i)}_u}$
  \State $\mathbf{P}^{(i)} \gets \argmin_{\mathbf{P} \in U(n_i,n_{\textrm{ref}})} \mathcal{L}_\lambda(\mu^{(i)}, \mu_{\textrm{ref}}, \mathbf{P})$
  \State $\mathbf{T}^{(i)} \gets (n_{\textrm{ref}} \mathbf{P}^{(i)})^T \mathbf{Y}^{(i)}$
  \State $\tilde{\mathbf{Y}}^{(i)} \gets (n_i \mathbf{P}^{(i)}) \mathbf{T}^{(i)}$
  \EndFor
  \State $\{\mathbf{C}^{(i)}\}_{i=1}^{N} \gets  \mathcal{M}_{DR}(\{\mathbf{T}^{(i)}\}_{i=1}^N)$\Comment{Reduced dimension $Q$}
  \For{$j=1, \cdots, Q$}
  \State $\mathcal{M}_{gp,j} \gets GP(\{ (X^{i}, C^{(i)}_j) \}^{N}_{i=1})$ 
  \EndFor
  \end{algorithmic}
\end{algorithm}

\begin{algorithm}[H]
  \caption{TOS-GP, test phase}\label{algo:tos-gp-test}
  \begin{algorithmic}[1]
  \Require{Test input $X_{*}$, reference measure $\mu_{\textrm{ref}}$, regularization parameter $\lambda$, (linear) dimension reduction model $\mathcal{M}_{DR}$, list of $Q$ GP models $\{ \mathcal{M}_{gp,j}\}_{j=1}^Q$} 
  \Ensure{Predicted signal $\hat{\mathbf{Y}}_{*}$, uncertainties $\{(\hat{\mathcal{S}}_{*})_{ii}^{\vphantom{ii}} \}_{i=1}^{n_{*}}$}
    %\Procedure{Euclid}{$a,b$}\Comment{The g.c.d. of a and b}
  \For{$j=1, \cdots, Q$}
  \State $\hat{C}_{*,j}, \hat{\sigma}_{*,j} \gets \mathcal{M}_{gp,j}(\{ (X^{(i)} \}_{i=1}^{N})) $
  \EndFor
  \State $\hat{\mathbf{C}}_{*} \gets (\hat{C}_{*,j})_{j=1}^Q$
  \State $\hat{\mathbf{T}}_{*} \gets  \mathcal{M}_{DR}^{-1}(\hat{\mathbf{C}}_{*})$\Comment{Transferred signal prediction}
  \State $\hat{\mathcal{S}}_{*,T} \gets \mathcal{M}_{DR}^{-1}(\{\hat{\sigma}_j\}_{j=1}^{Q} )$\Comment{Transferred signal UQ}
  \State $\mu_{*} \gets \frac{1}{n_{*}} \sum_{u=1}^{n_{*}} \delta_{\mathbf{F}_{*,u}}$
  \State $\mathbf{P}_{*} \gets \argmin_{\mathbf{P} \in U(n_{*},n_{\textrm{ref}})} \mathcal{L}_{\lambda_0}(\mu_{*}, \mu_{\textrm{ref}}, \mathbf{P})$
  \State $\hat{\mathbf{Y}}_{*} \gets (\mathbf{n}_{*}\mathbf{P}_{*})\hat{\mathbf{T}}_{*}$ \Comment{Signal prediction} 
  \State $\hat{\mathcal{S}}_{*} \gets (n_{*}\mathbf{\mathbf{P}}_{*})\hat{\mathcal{S}}_{*,T} (n_{*}\mathbf{\mathbf{P}}_{*})^T$ \Comment{Signal UQ}
  \end{algorithmic}
\end{algorithm}

\paragraph{Prediction.}
\label{section:prediction}
For a new test input $X_{*}$ with $n_{*}$ nodes, we first encode it as an empirical measure and compute the transport plan to the reference measure $\mathbf{\mathbf{P}}_{*}$. We then compute the single-output GP predictions to form the predicted coefficients vector $\hat{\mathbf{C}}_{*} = (\hat{C}_{*,j})_{j=1}^Q$. We apply the inverse of the dimension reduction method to get the predicted transferred signal $\hat{\mathbf{T}}_{*}= \mathcal{M}_{DR}^{-1}(\hat{\mathbf{C}}_{*})$ on the reference measure, which is transported back to estimate the predicted signal $\hat{\mathbf{Y}}_{*} = (\mathbf{n}_{*}\mathbf{P}_{*})\hat{\mathbf{T}}_{*}$. Prediction uncertainties can be obtained through sampling in the reduced space according to the GP posterior distribution, and then going backward through the exact same reconstruction stages. Note that in the particular case of a linear dimension reduction technique such as PCA, having access to the projection matrix $E^T$ yields an analytical formula for prediction uncertainties. Denoting $\sigma_{*,1}, \cdots, \sigma_{*,Q}$ the posterior standard deviation for all $Q$ PCA coefficients, the posterior distribution of the transferred signal is Gaussian with covariance given by $\hat{\mathcal{S}}_{*,T} = \mathbf{E} Diag(\sigma_{*,1}^2, \cdots, \sigma_{*,Q}^2)\mathbf{E}^T$ and similarly for the reconstructed signal $\hat{\mathcal{S}}_{*} = (n_{*}\mathbf{\mathbf{P}}_{*})\hat{\mathcal{S}}_{*,T} (n_{*}\mathbf{\mathbf{P}}_{*})^T$.

\paragraph{Reference measure and hyperparameters.}
There are several possible strategies for constructing the reference measure. One can take a barycenter of measures, or a uniform measure over a reference shape, typically the convex hull of the union of the supports of all train measures. The most natural and simplest option is to sub-sample from an original measure coming from the training set. A low discrepancy sequence is built from this empirical measure (using the Maximum Mean Discrepancy as described in the supplementary material) in order to reduce its support to a few representative points. The TOS-GP method also depends on several hyperparameters that need to be selected. When using PCA, the low-embedding dimension $Q$ can be chosen to achieve 95\% cumulative explained variance. Note, however, that limiting the number of PCA coefficients to smaller values is already sufficient for equivalent quality when reconstruction errors are already large. The regularization parameter $\lambda$ as well as the number of continuous WL iterations can be chosen either in a supervised manner or independently of training, so as to minimize the reconstruction error on the train samples only. Let us emphasize, however, that the number of continuous WL iterations and the choice of the reference measure are far less influential than regularization, as illustrated in our experiments.

\section{EXPERIMENTS} \label{section:exp}

For our numerical experiments \footnote{Code: \url{https://gitlab.com/drti/tos_gp/}.}, we focus on regression tasks involving large graphs from mesh-based simulations in computational fluid dynamics and mechanics \footnote{Datasets: \url{https://plaid-lib.readthedocs.io/en/latest/source/data_challenges.html}.}. 
TOS-GP is compared to three competing approaches: Mesh Morphing Gaussian Process (MMGP) \citep{mmgp} and two state-of-the-art GNN architectures: GCNN \citep{gnn_gcnn_kipf} and MGN \citep{gnn_mgn_pfaff2020}. In particular, we perform an in-depth study of the impact of various hyperparameters: the choice of the reference measure, the number of WL iterations and the regularization parameter.

\begin{figure}[H]
\centering
\includegraphics[scale=0.42]{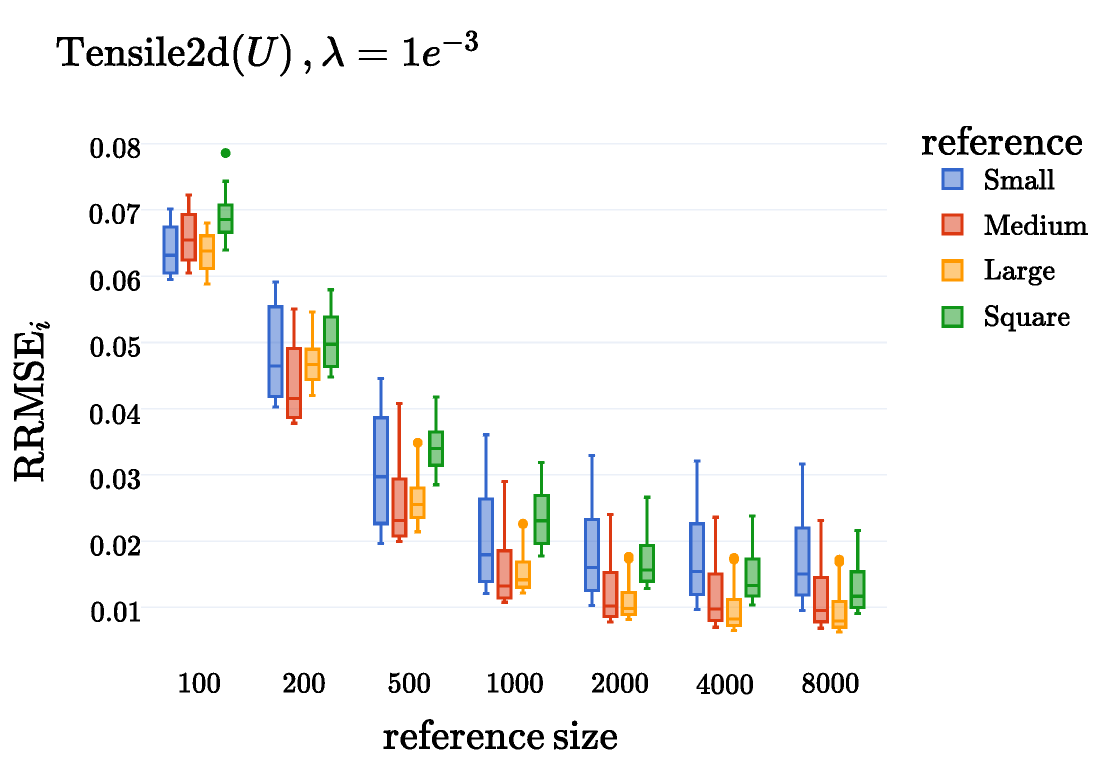}
\includegraphics[scale=0.42]{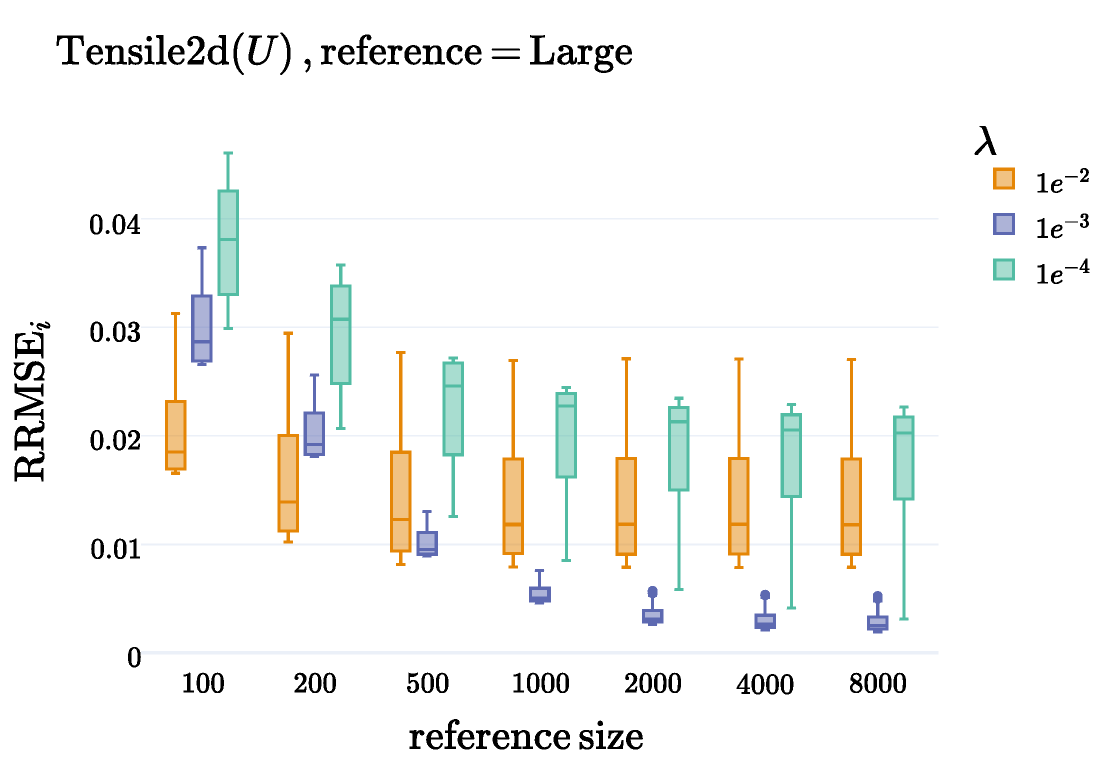}
\caption{\texttt{Tensile2} field \texttt{U}. Top: RRMSE vs reference size and reference measure with fixed $\lambda=1e-3$. Bottom: RRMSE vs reference size and $\lambda$ for a fixed reference measure.}
\label{fig:Tensile2d_evolution}
\end{figure}

For all considered problems, we use PCA as a dimension reduction technique, and choose the number of coefficients so as to obtain 95\% explained variance. The quality of trained models is assessed by computing the Relative Root Mean Square Error (RRMSE) following \cite{mmgp}. 

The RRMSE is defined for $N_*$ ground truth test signals $\{ \mathbf{Y}^{(i)}\}_{i=1}^{N_*}$, and the predictions $\{ \hat{\mathbf{Y}}^{(i)}\}_{i=1}^{N_*}$ (where the signal $i$ is of size $n_{*i}$) as:
\begin{equation}
\label{eq:RRMSE}
\RRMSE\left( \{ \mathbf{Y}^{(i)}\}_{i=1}^{N_*},  \{ \hat{\mathbf{Y}}^{(i)}\}_{i=1}^{N_*} \right)^2 = \frac{1}{N_*} \sum_{i=1}^{N_*}  \RRMSE_i^2 ,
\end{equation}
where $\RRMSE_i^2 =\frac{\| \mathbf{Y}^{(i)} - \hat{\mathbf{Y}}^{(i)}\|_2^2}{n_{*i} \|\mathbf{Y}^{(i)} \|^2_\infty}$  is the contribution of the $i$-th signals $Y^{(i)}$ and $\hat{Y}^{(i)}$ to $\RRMSE^2$. For each dataset, the methods are repeated $10$ times to test their variability with respect to their own random initializations, and we report the empirical mean and standard deviation of these 10 $\RRMSE$. Additional details on the GP model are given in the supplementary material.

\begin{table*}[b]
\setlength{\tabcolsep}{3pt}
\caption{Empirical means and standard deviations of the RRMSEs for all considered datasets and methods.}
\label{tab:regression}
\centering
\begin{tabular}{ccccccc} 
 \toprule
 Method\textbackslash Dataset & \texttt{Rotor37}(\texttt{P}) & \texttt{Rotor37}(\texttt{T}) & \texttt{Tensile2d}(\texttt{U}) & \texttt{Tensile2d}(\texttt{$\sigma_{12}$})  \\
 %& x10\textsuperscript{-3} & x10\textsuperscript{-3} & x1 & x1\\
 \midrule
 TOS-GP & 3.4e-2 (6e-4) & 9.6e-3 (2e-5) & 2.2e-3 (8e-6) & 5.6e-3 (3e-6)\\
 GCNN & 1.7e-2 (8e-4) & 3.9e-3 (1e-4) & 4.5e-2 (1e-2) & 4.5e-2 (4e-3) \\
 MGN & 1.7e-2 (2e-3) & 1.4e-2 (2e-3) & 1.5e-2 (1e-3) & 7.5e-3 (4e-4) \\
 MMGP & 7.2e-3 (5e-4) & 8.2e-4 (1e-5) & 3.4e-3 (4e-5) & 2.4e-3 (2e-5) \\
 \bottomrule
\end{tabular}
\end{table*}

\paragraph{Tensile2d dataset.}
We consider a two-dimensional problem in solid mechanics introduced by \cite{Tensile2dDataset}. The input geometries consist of 2D squares with two half circles that have been cut off in a symmetrical manner. These structures are subject to a uniform pressure field over the upper boundary and the material is modeled by a nonlinear elasto-viscoplastic law. The inputs of the problem are given by the mesh of the geometry (graphs with $\sim 10000$ nodes), and six scalar parameters that correspond to the material parameters and the input pressure applied to the upper boundary. The signal outputs of the problem are the horizontal component of the displacement field \texttt{U} and the tangential force per unit area acting in the horizontal direction on a surface normal to the second axis (shear stress \texttt{$\sigma_{12}$}) obtained with the \texttt{Zset} mechanical solver \cite{garaud2019z}. There are 500 train and 200 test samples.

We consider four types of references: \texttt{Small}, \texttt{Medium}, \texttt{Large} and \texttt{Square}. They are described in the supplementary material. For each type, we consider several support sizes ranging from 100 to 8000. We first study the influence of these hyperparameters in Figure \ref{fig:Tensile2d_evolution} for the \texttt{Tensile2d} field \texttt{U}. We notice that the RRMSE decreases as a function of the size of the reference measure, and seems to remain close to a constant beyond 1000 points. This suggests to choose the size of the reference measure as a tradeoff between computational time for OT and prediction errors. We also observe that the best regularization parameter is $\lambda = 1e-3$ which highlights the benefits of a well-chosen regularization. Interestingly, the \texttt{Large} reference measure with the largest surface gives slightly better scores with less variance than the \texttt{Medium} reference measure corresponding to the more central distribution. The \texttt{Square} reference gives equivalent performance when the size is high, but it seems to have more degraded predictions when the reference size is reduced compared to other references.

\smallskip

We also report in Table \ref{tab:regression} the results of TOS-GP with an exhaustive grid search to identify the best regularization parameter and number of continuous WL iterations to minimize the reconstruction error on the training set, the experimental results of the concurrent methods being taken from \cite{mmgp}. Remark that we achieve competitive scores compared to the state-of-the-art methods on the \texttt{Tensile2d} dataset.

\begin{figure}[h!]
    \centering
    \includegraphics[width=0.23\textwidth]{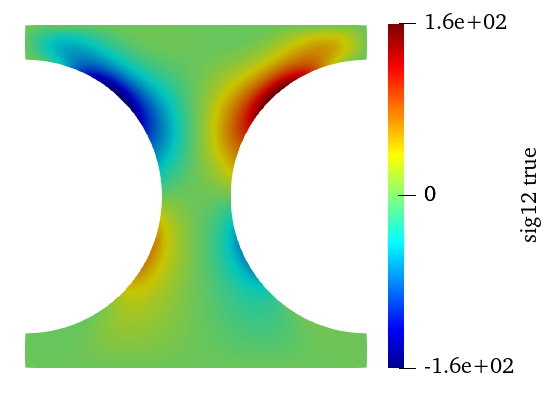}
    \includegraphics[width=0.23\textwidth]{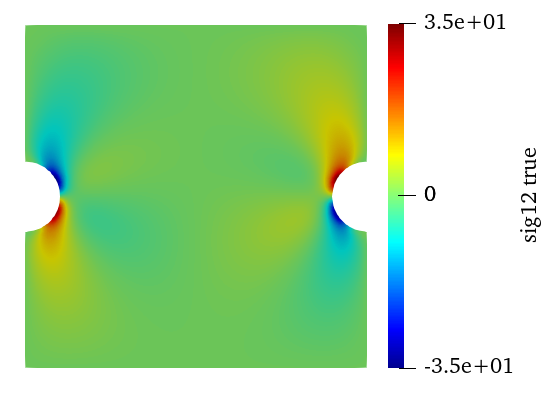}
    
    \includegraphics[width=0.23\textwidth]{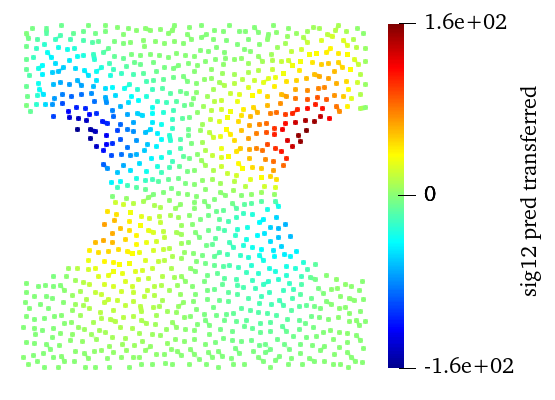}
    \includegraphics[width=0.23\textwidth]{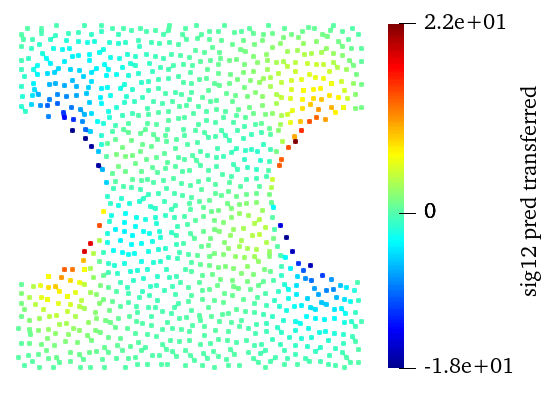}
    
    \includegraphics[width=0.23\textwidth]{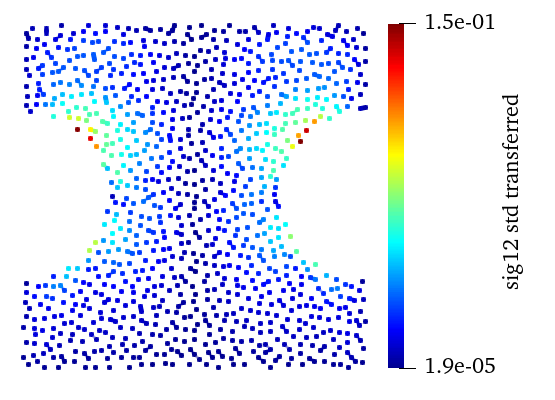}
    \includegraphics[width=0.23\textwidth]{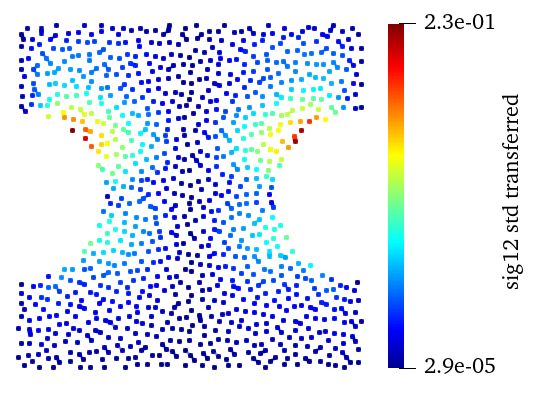}
        
    \includegraphics[width=0.23\textwidth]{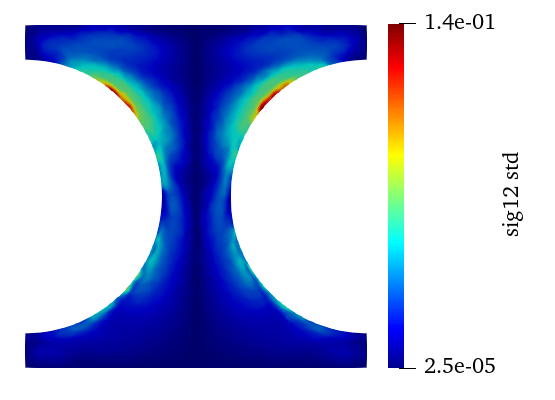}
    \includegraphics[width=0.23\textwidth]{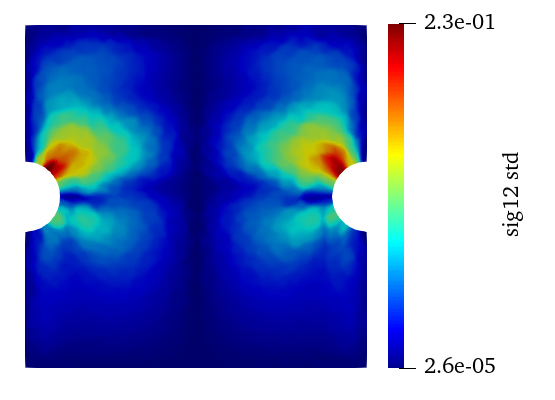}
    
    \caption{\texttt{Tensile2d}, \texttt{$\sigma_{12}$}: two test meshes (left and right) transported to a common reference. From top to bottom: output signals, predicted transferred signals, posterior standard deviation of the predicted transferred fields, posterior standard deviation of the predicted field.}
    \label{fig:tensile2d_fields_with_std}
\end{figure}

\paragraph{Rotor37 dataset.} The NASA rotor 37 case \citep{Rotor37Dataset} serves as a prominent example of a transonic axial-flow compressor rotor widely employed in computational fluid dynamics research. 
This dataset is made of 3D compressible steady-state Reynold-Averaged Navier-Stokes (RANS) simulations that model external flows \citep{ameri2009nasa}. The inputs are the FEM meshes of a 3D compressor blade (graphs with $\sim 30000$ nodes), as well as two additional scalar physical parameters corresponding to the rotational speed and the input pressure. The field outputs of the problem are the temperature \texttt{T} and the pressure \texttt{P} at each node. There are 1000 train and 200 test samples.

\begin{figure*}[ht]
    \centering
    
    \includegraphics[width=0.24\textwidth]{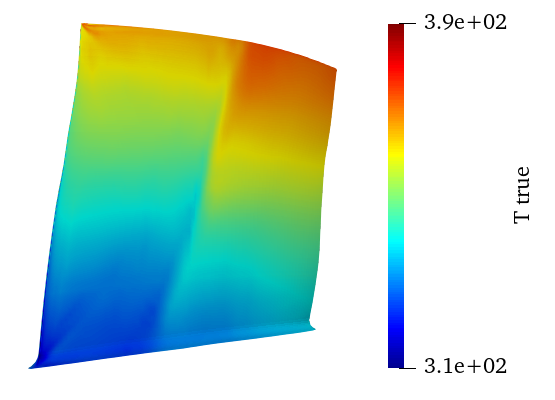}
    \includegraphics[width=0.24\textwidth]{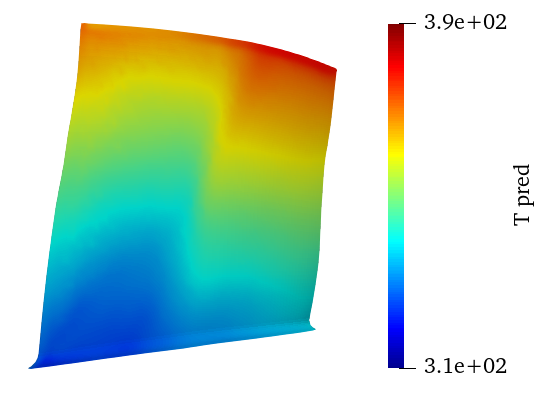}
    \includegraphics[width=0.24\textwidth]{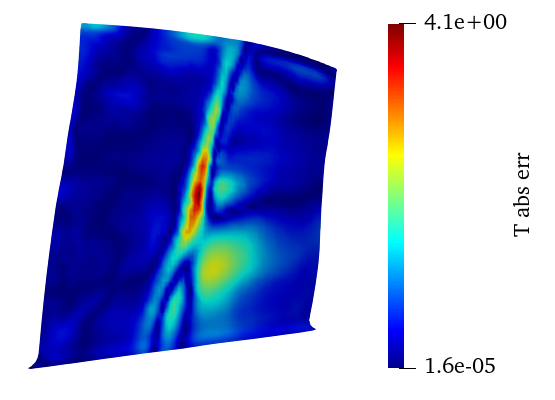}
    \includegraphics[width=0.24\textwidth]{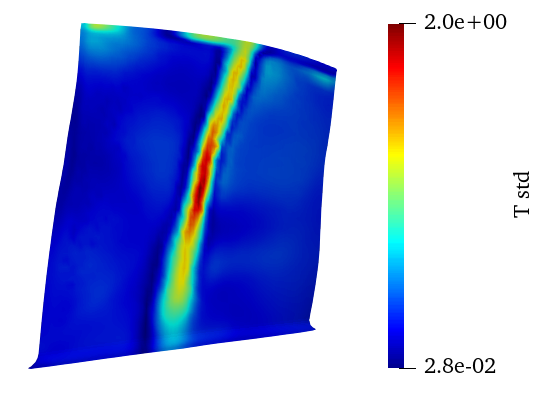}

    \caption{\texttt{Rotor37} test field \texttt{T} for one test input mesh. From left to right: true field, predicted field (posterior mean), absolute error, posterior standard deviation.}
    \label{fig:mosaic_fields}
\end{figure*}

Similarly to the previous test case, we consider both a reference measure coming from the training set and the circumscribed sphere around the training samples, see Table \ref{tab:regression} for the results obtained with the best set of hyperparameters. Here the RRMSE scores are comparable to those of GNN approaches but slightly lower than MMGP: this is due to the very strong discontinuity of the fields on the blade edge, which leads to a higher reconstruction error. Note that while the number of continuous WL iterations has little influence on \texttt{Tensile2d} outputs, it significantly helps with \texttt{Rotor37} output \texttt{P} as detailed in the supplementary material.
\vspace{-0.6mm}
\paragraph{Prediction uncertainty.}
A key benefit of TOS-GP is its ability to assess prediction uncertainties. In Figure \ref{fig:tensile2d_fields_with_std}, we use the common \texttt{Medium} reference of size 1000 to shed light on the uncertainty propagation described in Section \ref{section:prediction}. After applying the inverse of the dimension reduction method, both predictions and uncertainties are expressed on the same reference measure (second and third lines). Predictive uncertainties are then propagated to the original space (fourth line). We observe more spread-out uncertainties on the right-hand input mesh, which is actually more difficult to predict due to more localized constraints (its $\mathrm{RRMSE}_i$ is equal to $2.9e-2$ versus $1.1e-2$ for the left-hand input mesh). Figure \ref{fig:mosaic_fields} highlights predictions for \texttt{Roto37}'s field \texttt{T} along with their uncertainties and prediction errors. Interestingly, uncertainties are higher in ``critical" zones where the signal is likely to have the most variations. These zones notably correspond to the areas where the error is the biggest. More figures detailing uncertainties for other data sets are given in the supplementary.

\paragraph{Implementation and computing infrastructure.}
We leverage a Python implementation of GP regression with \citep{gpy}. The optimal transport plans are computed using the ott-jax library \citep{ott-jax}. All our analyses were performed on a hybrid (CPU/GPU) computational node using 1 Nvidia A100 GPU and 32 CPU cores (AMD EPYC Milan 7763) with a total of 128GB of RAM (4GB per core). 

\paragraph{Computation times.}
The most computationally intensive step of TOS-GP is the identification of the transport plans during the preprocessing step. Transfer times depend on both the size of the measures involved and the regularization parameter. For \texttt{Tensile2d} using a reference measure of size 8000, computing one transport plan takes 6 seconds for $\lambda=1e-3$. This corresponds to a total sequential preprocessing time of 2h38min for all training samples and all $\lambda$. Similarly, for \texttt{Rotor37} (temperature), computing one transport plan takes 45 seconds for a reference measure of size 15000 with $\lambda=1e-7$, and a total sequential time of 28h20min for all $\lambda$ and all training samples. But this preprocessing step is obviously embarrassingly parallel, reducing for example to 1min35s and 17min using 100 parallel jobs for \texttt{Tensile2d} and \texttt{Rotor37}, respectively. Computation times for concurrent methods are given in the supplementary material.

\section{INPUT GEOMETRIES WITH VARYING TOPOLOGIES}\label{section:varying_topologies}
In this section, we address a challenging regression problem involving topological changes in the input geometries. These topology variations can be handled by our methodology, which is not the case with MMGP. In addition, this dataset allows us to demonstrate the flexibility of the TOS-GP framework by considering different dimension reductions and kernels.

We introduce a two-dimensional multiscale problem in computational mechanics, where the goal is to solve a nonlinear boundary value problem to determine the macroscopic mechanical properties of a highly heterogeneous material at the microscopic scale as in \cite{pasparakis2024bayesian}. Four topologies are considered by generating microstructures with $5$ to $8$ pores. Mesh sizes range from 4119 to 7128 nodes. Examples of input meshes which represent microstructures of porous hyperelastic materials are shown in the supplementary material. The mechanical problem is also parameterized by three input scalars (components of the right Cauchy deformation tensor) which define the boundary conditions. Train and test datasets of respective sizes 762 and 376 are generated by a constrained design of experiments. We consider the horizontal component of the displacement field \texttt{U} as the output signal. 

The TOS-GP methodology is applied with a reference uniform grid of size $32\times 32$ on the unit square. A regularization parameter $\lambda=1e-3$ is used. For the Gaussian process regression, we consider a kernel that computes the MMD distance between the centers of the pores and plugs them in a squared exponential kernel. For the dimension reduction part, we observe that the problem is highly non-linear and therefore cannot be easily reduced to a low dimensional space with PCA. We instead rely on convolutional autoencoders \citep{Goodfellow-et-al-2016} built on the transferred fields defined on the $32\times 32$ unit grid. More specifically, the auto-encoder combines convolutional and 2D discrete Fourier layers based on the Fourier neural operator approach of \cite{fno_ae}, providing encodings in dimension $2\times 8\times 8=128$.

\begin{figure}[h!]
\centering
    
\includegraphics[width=0.23\textwidth]{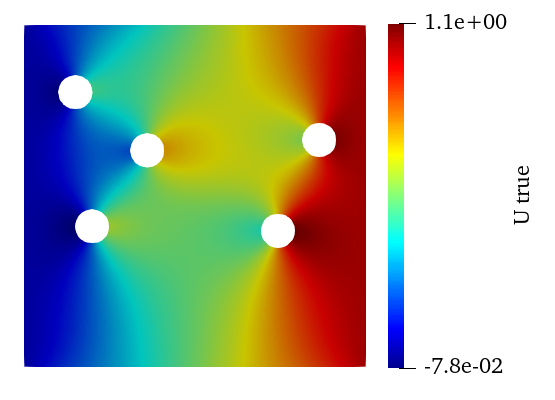}
\includegraphics[width=0.23\textwidth]{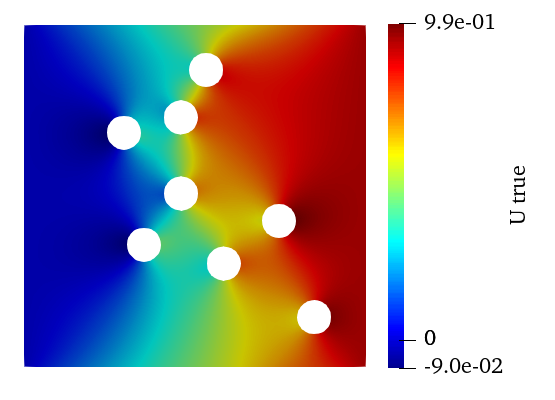}

\includegraphics[width=0.23\textwidth]{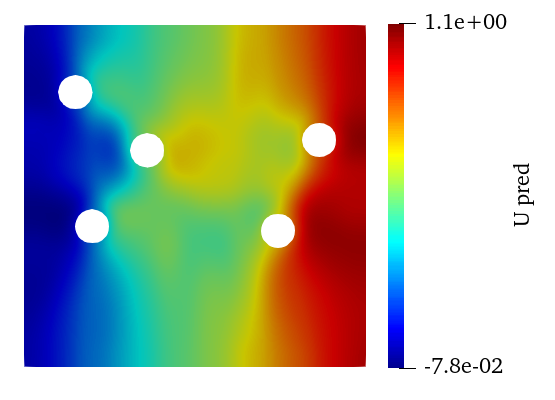}
\includegraphics[width=0.23\textwidth]{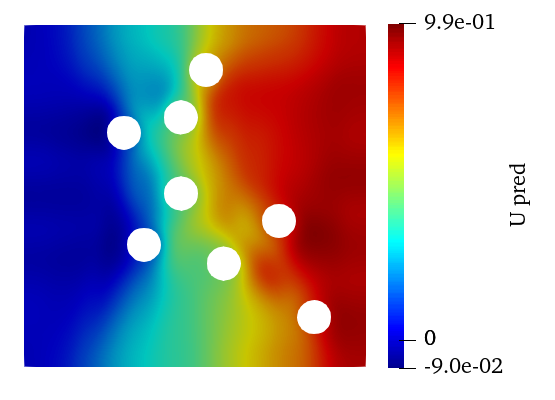}

\includegraphics[width=0.23\textwidth]{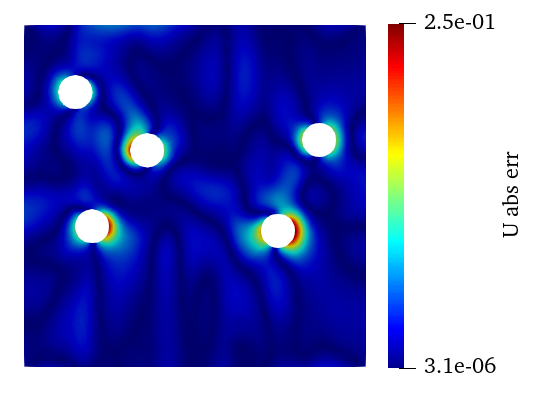}
\includegraphics[width=0.23\textwidth]{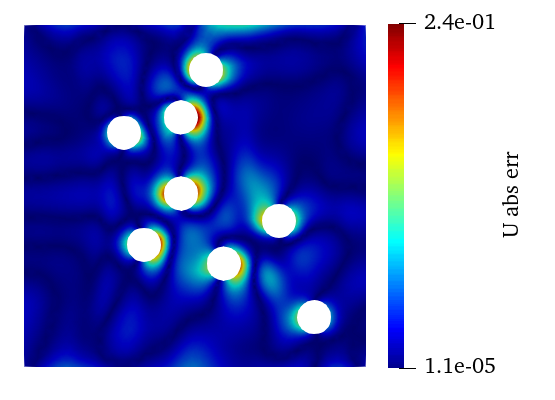}

\includegraphics[width=0.23\textwidth]{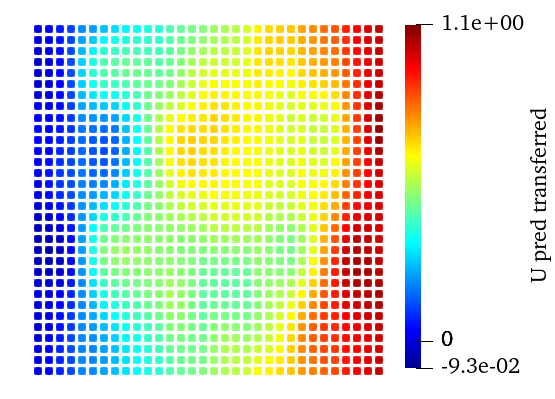}
\includegraphics[width=0.23\textwidth]{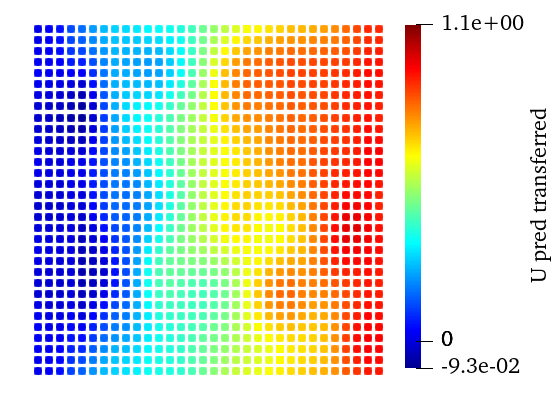}

    \caption{Two samples with different topologies from the multiscale dataset. From top to bottom: true field, predicted field, absolute error, predicted transferred signal.}
    \label{fig:mosaic_fields_emmentalisticity}
\end{figure}

Figure \ref{fig:mosaic_fields_emmentalisticity} shows predictions for two sample meshes with 5 and 8 pores respectively. Even though their topologies change, regularized optimal transport allows to express output signals on the reference grid, thus enabling the predictions in this transferred space (see the rightmost figures). Here, we demonstrate that our method can seamlessly handle input meshes with varying topologies without requiring any modifications. It is worth emphasizing that to our best knowledge, such problems have so far only been addressed by relying on deep neural networks. A comparison with MGN is included in the supplementary.

\section{CONCLUSION}
We introduce TOS-GP, a first GP-based supervised model that can predict outputs defined as fields discretized on large graphs with continuous node attributes. It demonstrates similar predictive capabilities in comparison with state-of-the-art GNN architectures for several problems in computational physics, while ensuring uncertainty quantification on the nodes and involving few hyperparameters to tune. We focused here on datasets coming from physical simulations. But a key feature is that, since we make no assumption on the mesh topology nor on the graph adjacency, our methodology can also be readily extended to predict signals defined on point clouds or on meshes with topology changes. In addition, since the OT and dimension reduction steps are agnostic to the choice of the supervised model, one can consider replacing GP with any other single-output ML model. To go further, we observe that the regularized OT between measures of different sizes leads to an approximation error that may be high, which in practice can be compensated by an appropriate choice of the regularization parameter. But signals with higher variability, whose regularity varies greatly between inputs, needs specific care. As a perspective, in order to better take adjacency into account, we plan to investigate transport plans using the Fused Gromov Wasserstein distance \citep{fgw,vincent2022template}, but this will certainly come with a significant computation overhead. Lastly, while our approach decouples output transfer and learning, one could also design variants of the work of \cite{bachoc2020gaussian} and \cite{bachoc2023gaussian} that reuse the same transport plans in both input and output spaces.

%\subsubsection*{Acknowledgements}
%This work was partially supported by the Agence Nationale de la Recherche through the SAMOURAI (Simulation Analytics and Metamodel-based solutions for Optimization, Uncertainty and Reliability AnalysIs) project under grant ANR20-CE46-0013 and the EXAMA (Methods and Algorithms at Exascale) project under grant ANR-22-EXNU-0002.

%\subsubsection*{References}
\bibliography{b}

\section*{Checklist}

 \begin{enumerate}

 \item For all models and algorithms presented, check if you include:
 \begin{enumerate}
   \item A clear description of the mathematical setting, assumptions, algorithm, and/or model. [Yes]
   \item An analysis of the properties and complexity (time, space, sample size) of any algorithm. [Yes]
   \item (Optional) Anonymized source code, with specification of all dependencies, including external libraries. [Yes]
 \end{enumerate}

 \item For any theoretical claim, check if you include:
 \begin{enumerate}
   \item Statements of the full set of assumptions of all theoretical results. [Not Applicable]
   \item Complete proofs of all theoretical results. [Not Applicable]
   \item Clear explanations of any assumptions. [Not Applicable]     
 \end{enumerate}

 \item For all figures and tables that present empirical results, check if you include:
 \begin{enumerate}
   \item The code, data, and instructions needed to reproduce the main experimental results (either in the supplemental material or as a URL). [Yes]
   \item All the training details (e.g., data splits, hyperparameters, how they were chosen). [Yes]
         \item A clear definition of the specific measure or statistics and error bars (e.g., with respect to the random seed after running experiments multiple times). [Yes]
         \item A description of the computing infrastructure used. (e.g., type of GPUs, internal cluster, or cloud provider). [Yes]
 \end{enumerate}

 \item If you are using existing assets (e.g., code, data, models) or curating/releasing new assets, check if you include:
 \begin{enumerate}
   \item Citations of the creator If your work uses existing assets. [Yes]
   \item The license information of the assets, if applicable. [Yes]
   \item New assets either in the supplemental material or as a URL, if applicable. [Yes]
   \item Information about consent from data providers/curators. [Not Applicable]
   \item Discussion of sensible content if applicable, e.g., personally identifiable information or offensive content. [Not Applicable]
 \end{enumerate}

 \item If you used crowdsourcing or conducted research with human subjects, check if you include:
 \begin{enumerate}
   \item The full text of instructions given to participants and screenshots. [Not Applicable]
   \item Descriptions of potential participant risks, with links to Institutional Review Board (IRB) approvals if applicable. [Not Applicable]
   \item The estimated hourly wage paid to participants and the total amount spent on participant compensation. [Not Applicable]
 \end{enumerate}

 \end{enumerate}

%%%%%%%%%%%%%%%%%%%%%%%%%%%%%%%%%%%%%%%%%%%%%%%%%%%%%%%%%%%%

\appendix
\onecolumn
\aistatstitle{Supplementary Material for Learning signals defined on graphs with optimal transport and
Gaussian process regression}

%\vspace{-6\baselineskip}

\section{More experimental details}
This section provides more information about the experiments described in the main paper. We summarize the datasets used for the experiments. We give more details on the Gaussian process model, on the choice of the reference measures and on the hyperparameter selection. We also provide a more precise analysis of the preprocessing and training times.

\subsection{Datasets}
Details about the datasets are given in Table \ref{tab:data}. The attributes column corresponds to the dimension of the continuous attributes associated to each node. 
\begin{table*}[h]
\caption{Summary of the datasets. $(*)$: fixed number of nodes and adjacency structure}
\label{tab:data}
\centering
\begin{tabular}{ccccccc} 
 \toprule
  Dataset & Train+Test & Mean number & Mean number & Number of & Number of & Number of\\
   & samples & of nodes &  of edges & attributes  & input scalars & output fields\\
 \midrule
 Rotor37$ ^*$ & 1000+200 & 29773 & 77984 & 3 & 2 & 2 \\
 Tensile2d & 500+200 & 9425.6 & 27813.8 & 2 & 6 & 2 \\
 Multiscale & 764+376 & 4591.6 & 16721.6 & 2 & 3 & 1 \\
 %AirfRANS & 800+200 & 179779.0 & 536826.6 & 2 & 2 & Regression \\
 \bottomrule
\end{tabular}
\end{table*}

\begin{figure}[h]
\begin{center}
%\framebox[4.0in]{$\;$}
%\fbox{\rule[-.5cm]{0cm}{4cm} \rule[-.5cm]{4cm}{0cm}}
\includegraphics[width=0.24\textwidth]{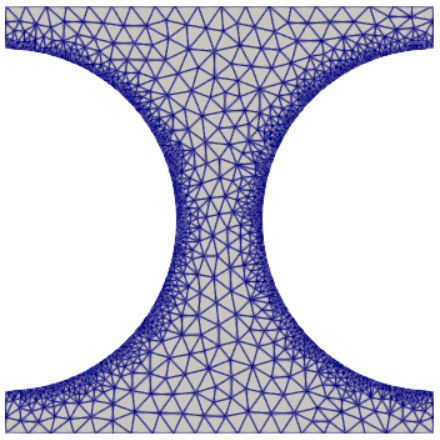}
\includegraphics[width=0.24\textwidth]{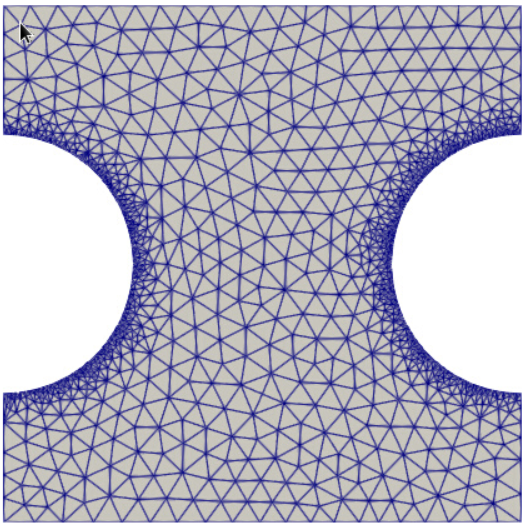}
\includegraphics[width=0.24\textwidth]{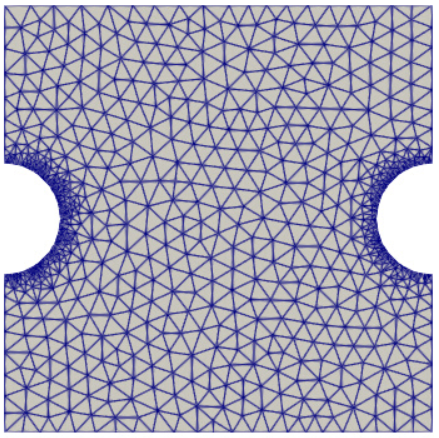}
\end{center}
\caption{\texttt{Small}, \texttt{Medium} and \texttt{Large} input meshes from the \texttt{Tensile2d} dataset (from left to right). The meshes are coarsened here for visualization purpose only.}
\label{fig:meshes}
\end{figure}

For \texttt{Tensile2d}, we consider three references in the train set respectively called \texttt{Small}, \texttt{Medium} and \texttt{Large} with original size 6143, 9733 and 11627, respectively. These \texttt{Small}, \texttt{Medium}, \texttt{Large} inputs actually correspond to the largest, medium-size and smallest radius of the half circles as shown in Figure \ref{fig:meshes}. Such measures are then sub-sampled to obtain new supports of pre-defined sizes included in the original supports. 

\subsection{Gaussian Processes}
The GP kernel function is chosen as the tensorized kernel
\begin{equation} \label{eq:tensorized}
     k(\mathfrak{X}, \mathfrak{X}') := \sigma^2 k_{\mathrm{SWWL}}(X,X') \prod_{\ell=1}^{m} c_{5/2}(|s_\ell-s'_\ell|)
\end{equation}
where $\mathfrak{X} = (X, s_1,\ldots, s_m)$ and $\mathfrak{X}' = (X', s'_1,\ldots, s'_m)$ denote the input graphs and the $m$ input scalars $s_1,\ldots,s_m$, $c_{5/2}$ is the Matérn-$5/2$ covariance function, $\sigma^2$ is a variance parameter and $k_{\mathrm{SWWL}}$ is a SWWL kernel with 50 projections and 500 quantiles. The lengthscale parameters of the SWWL and Matérn-5/2 kernels are optimized simultaneously by maximizing the marginal log-likelihood \citep{gpml}.

\subsection{Choice of the reference measure}
The reference measure can be chosen in various ways: selecting an arbitrary measure within the training dataset or computing a barycenter, for instance. In our work, we build a reference measure by subsampling a selected measure in the training dataset using a procedure described in the sequel of this section.
% We first present the procedure employed to obtain subsampled measures from a given empirical measure before discussing other ways to choose the reference measure.
Let $X$ be an arbitrary graph with node features $\mathbf{F}_1, \dots, \mathbf{F}_{n}$. 
% Let's choose an input $X^{(0)}$ with associated measure 
Let then $\mu = \frac{1}{n} \sum_{u=1}^{n} \delta_{\mathbf{F}_u}$ be the associated empirical measure. We subsample the measure $\mu$ using a procedure that minimizes the Maximum Mean Discrepancy (MMD) \cite{mmd1}.

\begin{definition}[Maximum Mean Discrepancy]
Let $x$ and $y$ be random variables defined on a topological space $\mathcal{Z}$, with respective Borel probability measures $p$ and $q$. 
% Let $\mathcal{F}$ be a class of functions $f: \mathcal{Z} \rightarrow \mathbb{R}$. 
Let $k : \mathcal{Z} \times \mathcal{Z} \rightarrow \mathbb{R}$ be a kernel function and let $\mathcal{H}(k)$ be the associated reproducing kernel Hilbert space. The maximum mean discrepancy between $p$ and $q$ is defined as
% \begin{equation}
% \label{eq:mmd}
%     \mathrm{MMD}(\mathcal{F}, p, q) = \sup_{f\in \mathcal{F}}\left( \mathbb{E}_{x\sim p}[f(x)] - \mathbb{E}_{y\sim p}[f(y)] \right)\,.
% \end{equation}
\begin{align*}
    \mathrm{MMD}_k(p, q) = \sup_{\|f\|_{\mathcal{H}(k)} \leq 1} |\mathbb{E}_{x \sim p}[f(x)] - \mathbb{E}_{y\sim q}[f(y)]|\,.
\end{align*}
\end{definition}
% The MMD has a closed-form expression when $\mathcal{F}$ is chosen to be the unit ball in a reproducing kernel Hilbert space $\mathcal{H}$. If we consider the positive definite kernel function $k: \mathcal{Z}\times \mathcal{Z} \rightarrow \mathbb{R}$, then we can write the squared population MMD as:\\
The MMD admits the following closed-form expression:
% \begin{equation}
%     \mathrm{MMD}(\mathcal{H}, p,q)^2 = \mathbb{E}_{x\sim p,x' \sim p}[k(x,x')] + \mathbb{E}_{y \sim q,y' \sim q}[k(y,y')] - 2\mathbb{E}_{x \sim p,y \sim q}[k(x,y)]\,,
% \end{equation}
\begin{equation}
    \mathrm{MMD}_k(p,q)^2 = \mathbb{E}_{x\sim p,x' \sim p}[k(x,x')] + \mathbb{E}_{y \sim q,y' \sim q}[k(y,y')] - 2\mathbb{E}_{x \sim p,y \sim q}[k(x,y)]\,,
\end{equation}
which can be estimated thanks to U- or V-statistics.
% where $x$ and $x'$ are independent random variables with distribution $p$, and $y$ and $y'$ are independent random variables with distribution $q$. 
% We can write $\mathrm{MMD}(\mathcal{H}, p,q) := \mathrm{MMD}_k(p,q)$. 
Here $k$ is chosen as the distance-induced kernel $k(x, y) = ||x|| + ||y|| - ||x - y||$ which has been shown to be characteristic by \cite{sejdinovic2013equivalence}.

% The MMD-subsampling algorithm \ref{algo:mmd} is the following: at each step, we maintain a set of points representing the empirical distribution, and we choose to add a new point that will minimize the MMD between the original measure and the subsampled one.
The MMD subsampling algorithm \ref{algo:mmd} selects $m < n$ particles by greedily minimizing the MMD. At the $(i+1)$-th iteration, a new particle $F_j$ is selected by minimizing the MMD between $\mu$ and the empirical measure given by 
\begin{align*}
    \mu_{i+1}(j) = \frac{1}{i + 1} \sum_{\ell \in \mathcal{P}_i} \delta_{\mathbf{F}_\ell} + \frac{1}{i + 1} \delta_{\mathbf{F}_j}\,, \quad j = 1,\dots,n\,,
\end{align*}
where $\mathcal{P}_i \subset \{1, \cdots,n\}$ denotes the set of particles indices already selected by the previous $i$ iterations.
% \begin{algorithm}
%   \caption{MMD subsampling}
%   \label{algo:mmd}
%   \begin{algorithmic}[1]
%   \Require{Empirical measure  $\mu_{0}$, kernel $k$, new size $S$}
%   \Ensure{Subsampled measure $\mu'_{0}$}
%   \State $\mathcal{P} \gets \varnothing$
%   \For{$i=1, \cdots, S$}
%   \State $j_i \gets \argmin_{j=1\cdots n_0} \mathrm{MMD}^2(\mu_0, \frac{1}{|\mathcal{P}|} \sum_{j\in \mathcal{P}}  \delta_{\mathbf{F}^{(0)}_j})$
%   \State $\mathcal{P} \gets \mathcal{P} \cup \{ j_i\}$
%   \EndFor
%   \State $\mu'_{0} \gets \frac{1}{S} \sum_{j\in \mathcal{P}}  \delta_{\mathbf{F}^{(0)}_j}$
%   \end{algorithmic}
% \end{algorithm}
\begin{algorithm}
  \caption{MMD subsampling}
  \label{algo:mmd}
  \begin{algorithmic}[1]
  \Require{Empirical measure  $\mu$, kernel $k$, subsample size $m$}
  \Ensure{Subsampled measure $\mu'$}
  \State $\pi_{1} \gets \argmin_{j\in\{1, \cdots, n\}} \mathrm{MMD}_k^2(\mu, \delta_{\mathbf{F}_j})$
  \State $\mathcal{P}_1  = \{ \pi_1 \}$
  \For{$i=1, \cdots, m-1$}
  \State $\pi_{i+1} \gets \argmin_{j\in\{1, \cdots, n\}\smallsetminus \mathcal{P}_{i}} \mathrm{MMD}_k^2(\mu, \mu_{i+1}(j))$
  \State $\mathcal{P}_{i+1} \gets \mathcal{P}_i \cup \{ \pi_{i+1} \}$
  \EndFor
  \State $\mu' \gets \frac{1}{m} \sum_{j\in \mathcal{P}_m}  \delta_{\mathbf{F}_j}$
  \end{algorithmic}
\end{algorithm}

When the input measures lie in a two-dimensional space, we can also choose a shape corresponding to the convex hull of the supports of all train measures and then choose a uniform distribution of suitable size  on it. This is less obvious in dimension three when the measure is the discretization of a 2-manifold. For \texttt{Rotor37}, we can use a circumscribed sphere to the union of all the supports of train measures. In practice, the scores obtained with such a reference are very similar to those obtained with a reference measure from the train. It is further possible to consider an optimal transport barycenter of the input measures as a reference. For \texttt{Tensile2d}, we observe that the barycenter is close to the input measure \texttt{Medium} already considered. For \texttt{Rotor37}, the barycenter no longer corresponds to the discretization of a 2-manifold, which is less relevant than choosing a train measure as a reference. Instead, it is possible to select a "representative" measure from the set of train measures. This can be done using MMD again, but this time in a graph space. The MMD subsampling algorithm \ref{algo:mmd} is applied to the set of train graphs using the SWWL kernel (with lengthscales equal to the median of pairwise SWWL pseudo-distances), with only one step to identify a central train input in the sense of SWWL pseudo-distances. In practice, we again observe few differences when taking the first input or the latter one as reference measures due to the proximity of the geometries.

\subsection{Hyperparameter selection}
In addition to the choice of the reference measure, the method involves various hyperparameters such as the number of continuous WL iterations, the regularization parameter and the number of PCA coefficients. 
For \texttt{Tensile2d}, we obtain subsampled versions of the \texttt{Small}, \texttt{Medium} and \texttt{Large} references with 100, 200, 500, 1000, 2000, 4000, 8000 points each with the MMD procedure described previously.  A uniform distribution on the convex hull of all inputs (a square) with the same sizes is also considered.
 We vary the regularization parameter $\lambda$ in $\{1e-4, 1e-3, 1e-2\}$. For \texttt{Rotor37}, we consider the first input measure in the dataset as a reference with 100, 1000, 5000, 10000 and 15000 points. Some preliminary tests with other references showed that it was not relevant to change it with another train input. The regularization parameter $\lambda$ is selected in $\{1e-7, 1e-6, 1e-5, 1e-4\}$ for the temperature field and in $\{1e-8, 1e-7, 1e-6\}$ for the pressure. The number of continuous WL iterations is selected in $\{0,1,2,3\}$ for all the problems. 

The number of PCA coefficients is chosen so as to obtain a cumulative variance ratio greater than 95\%. If this number is smaller than 4, we decide to use a minimum number of coefficients equal to 4. In practice, we thus need to learn 4 coefficients for both fields of \texttt{Tensile2d} and \texttt{Rotor37}(\texttt{T}), and 51 coefficients for \texttt{Rotor37}(\texttt{P}). Remark that taking only 10 coefficients for \texttt{Rotor37}(\texttt{P}) gives a similar prediction error between test and predicted transferred fields due to an already high approximation error coming from the optimal transport part. In the case of Tensile2d, we add a warm-start initialization of the lengthscales, variances and nuggets based on a common GP trained to learn all coefficients simultaneously in order to help find a better optimum of the likelihood function.

Results given in Table \ref{tab:regression} correspond to the following parameters:\\
- \texttt{Tensile2d}(\texttt{U}): \texttt{Large} reference measure with size 8000, $\lambda=1e-3$, 1 WL iteration, 4 PCA coefficients\\
- \texttt{Tensile2d}($\sigma_{12}$): \texttt{Large} reference measure with size 8000, $\lambda=1e-3$, 2 WL iterations, 4 PCA coefficients\\
- \texttt{Rotor37}(\texttt{T}): reference measure with size 15000, $\lambda=1e-6$, 1 WL iteration, 10 PCA coefficients\\
- \texttt{Rotor37}(\texttt{P}): reference measure with size 15000, $\lambda=1e-8$, no WL iteration, 10 PCA coefficients

\subsection{Detailed computation times}
The most costly steps are the obtention of transport plans during the preprocessing step and the learning phase with Gaussian processes. Transfer times depend on both the size of the measures involved and the regularization parameter. For \texttt{Tensile2d} using the \texttt{Large} reference measure with size 8000, computing optimal transport plans takes 3, 6 and 10 seconds for $\lambda=1e-2$, $\lambda=1e-3$ and $\lambda=1e-4$ respectively. This corresponds to a total sequential preprocessing time of 9500 seconds (2h38min) for all regularizations, which can be easily reduced by computing transport plans in parallel (in practice, only 1min35s for the entire train dataset with 100 parallel jobs). Considering all measure types (four choices), sizes (four choices with supports of respective sizes 100, 200, 500, 1000, 2000, 4000, and 8000) and regularizations (three choices) involved in Figure \ref{fig:Tensile2d_evolution}, this corresponds to a total sequential time of 142830 seconds (24 minutes using 100 parallel jobs). Surprisingly, changing the number of continuous WL iterations (and therefore the size of the points forming the input and output empirical measures) has a negligible impact on transport plan computation time. Remark that for the test phase, a preprocessing step to obtain the test transport plans is also necessary. It takes 1200 seconds in total for the 200 test outputs (again possible to parallelize to be broken down to a 12 seconds using 100 jobs). Gaussian processes take 8min40s to train (with 3 restarts). 

\pagebreak
Similarly, for \texttt{Rotor37}, transport plans take 1, 4, 16, 34 and 52 seconds when the reference is the first input being subampled to have respective support size equal to 100, 1000, 5000, 10000 and 15000 with $\lambda=1e-8$. Fixing the support size to 15000, transport plans take respectively 52, 45, 40, 6 and 2 seconds for $\lambda = 1e-8, 1e-7, 1e-6, 1e-5, 1e-4$. For the pressure field, considering all sizes (100, 1000, 5000, 10000 and 15000) and regularizations (three choices, $\lambda = 1e-8, 1e-7, 1e-6$), this corresponds to a total sequential time of 316430 seconds (53 minutes using 100 parallel jobs). For the temperature field, considering all sizes (100, 1000, 5000, 10000 and 15000) and regularizations (four choices, $\lambda = 1e-7, 1e-6, 1e-5, 1e-4$), this corresponds to a total sequential time of 206127 seconds (34 minutes using 100 parallel jobs). Gaussian processes take 2h to train (with 3 restarts) for the field \texttt{P} and 11min for the field \texttt{T}. In comparison, GCNN takes respectively 24h and 1h for \texttt{Rotor37} and \texttt{Tensile2d} and MGN takes respectively 13h and 7h for \texttt{Rotor37} and \texttt{Tensile2d} for each hyperparameter set according to \cite{mmgp}.

\section{More experimental results}
In this section, we add experimental results. In particular, we present visuals of the predicted fields for all the datasets in Section \ref{section:exp} in addition to those of the main paper. We also give details about the errors, and the influence of the number of continuous WL iterations. Lastly, we present an additional comparison with MGN for the multiscale hyperelasticity problem of Section \ref{section:varying_topologies}.

\subsection{Predicted fields and uncertainties}
\begin{figure*}[ht]
    \centering
    \includegraphics[width=0.24\textwidth]{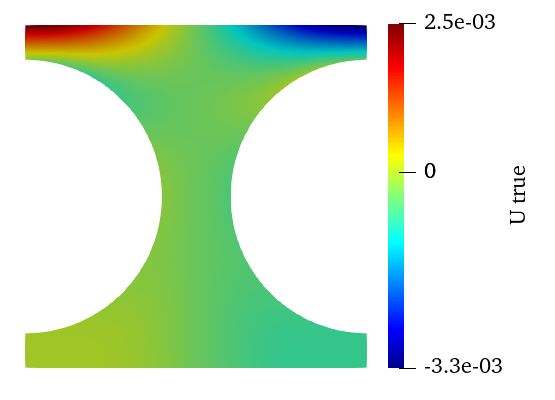}
    \includegraphics[width=0.24\textwidth]{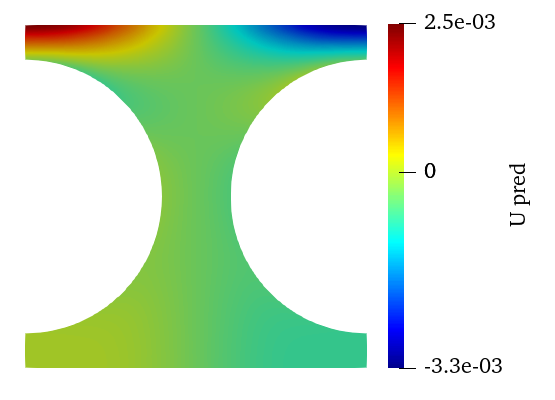}
    \includegraphics[width=0.24\textwidth]{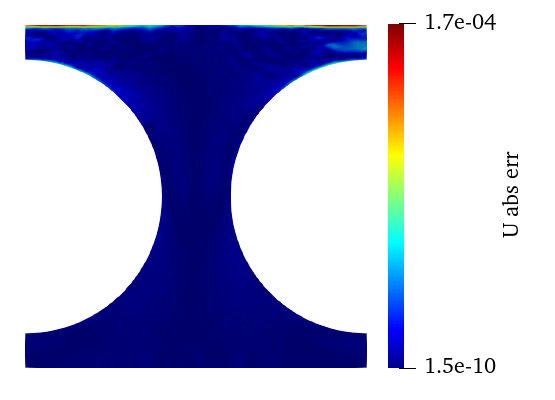}
    \includegraphics[width=0.24\textwidth]{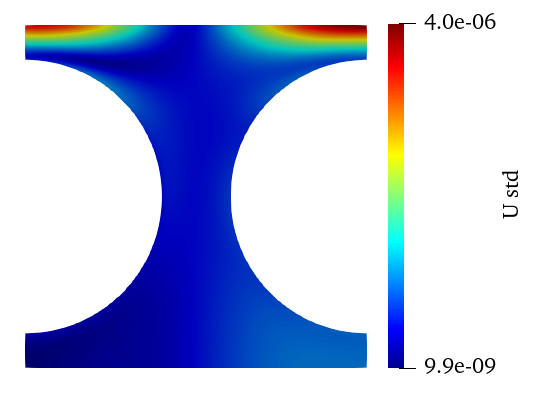}
    
    \includegraphics[width=0.24\textwidth]{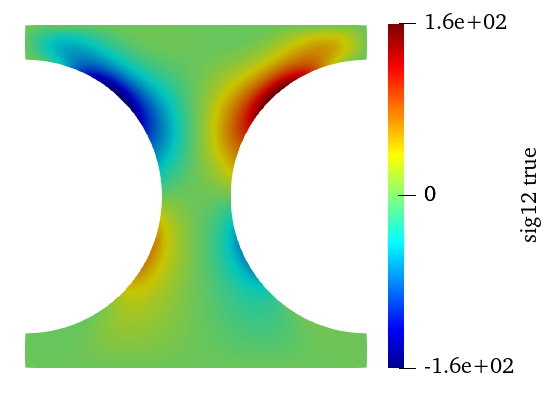}
    \includegraphics[width=0.24\textwidth]{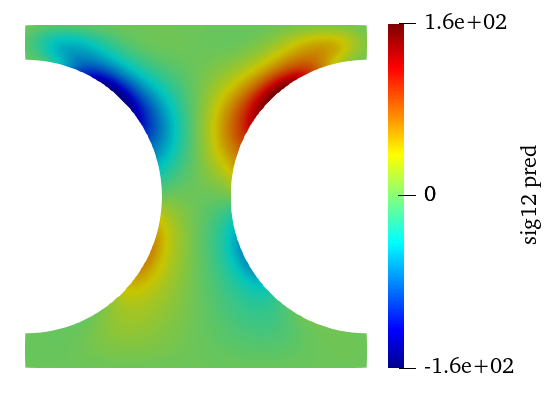}
    \includegraphics[width=0.24\textwidth]{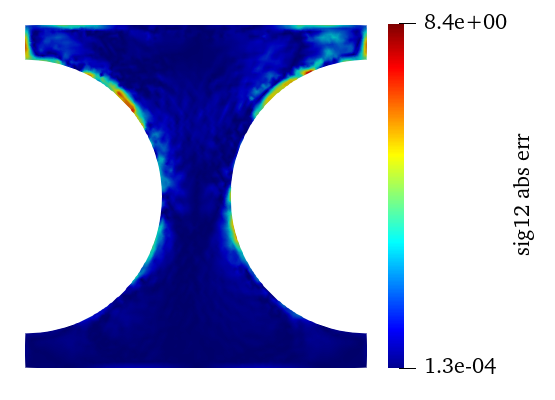}
    \includegraphics[width=0.24\textwidth]{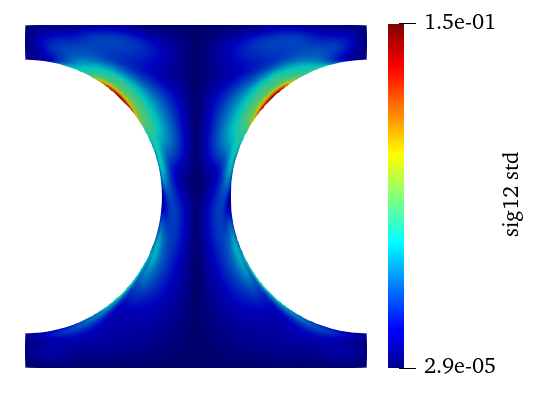}

    \includegraphics[width=0.24\textwidth]{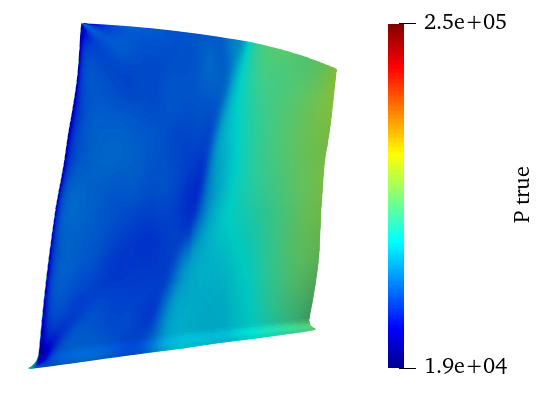}
    \includegraphics[width=0.24\textwidth]{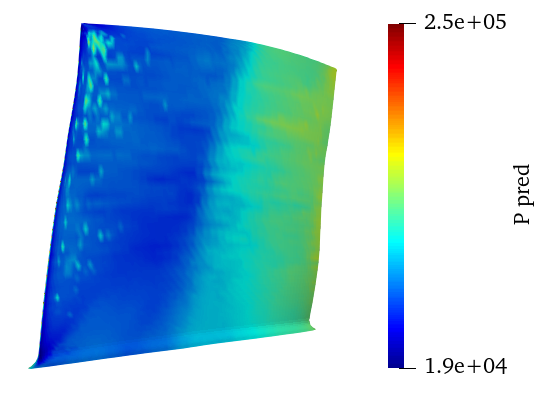}
    \includegraphics[width=0.24\textwidth]{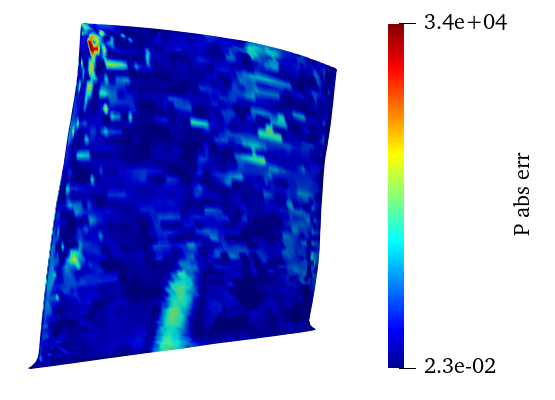}
    \includegraphics[width=0.24\textwidth]{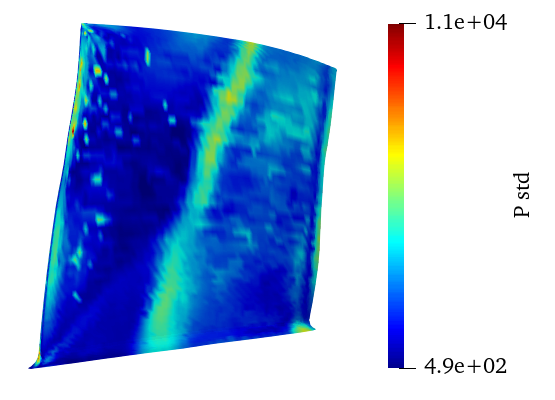}
    
    \caption{From top to bottom: \texttt{Tensile2d} fields \texttt{U} and \texttt{T}, \texttt{Rotor37} field \texttt{P} for one test input mesh. From left to right: true field, predicted field, absolute error, standard deviation from the posterior law.}
    \label{fig:mosaic_fields_bonus}
\end{figure*}

Figure \ref{fig:mosaic_fields_bonus} displays the predictions and uncertainties for the datasets of Section \ref{section:exp}. Similarly to Figure \ref{fig:mosaic_fields} , uncertainties are concentrated in areas where the signal varies the most and also seem to correspond to high error zones. For \texttt{Rotor37}(\texttt{P}), we observe a strange behavior with the prediction, which shows small areas on the left. This is due to a very large signal discontinuity on the slice of the mesh (with very high signal values), which is sent to the face via the regularized optimal transport plan. Unfortunately, this error is found as early as the transfer phase, and taking smaller regularizations does not improve this transfer error. 
Both Figure \ref{fig:mosaic_fields} and Figure \ref{fig:mosaic_fields_bonus} represent the first test input of the \texttt{Rotor37} dataset. Figure \ref{fig:mosaic_fields_bonus} and Figure \ref{fig:tensile2d_fields_with_std} respectively use the test inputs with the largest and smallest radius of the half circles.

\subsection{Prediction errors}

The detail of the prediction error due to the reconstruction and due to the Gaussian process model is given in Table \ref{tab:detail_errors}. The approximation error $\RRMSE\left( \{ \mathbf{Y}^{(i)}\}_{i=1}^{N_*},  \{ \tilde{\mathbf{Y}}^{(i)}\}_{i=1}^{N_*} \right)$ corresponds to the error between true and approximated signals (that is to say the signal obtained by transferring the true field, and then transferring it back without any prediction) while the transferred prediction error $\RRMSE\left( \{ \mathbf{T}^{(i)}\}_{i=1}^{N_*},  \{ \hat{\mathbf{T}}^{(i)}\}_{i=1}^{N_*} \right)$ corresponds to the error between the true and predicted signals in the transferred space. Note that the total error $\RRMSE\left( \{ \mathbf{Y}^{(i)}\}_{i=1}^{N_*},  \{ \hat{\mathbf{Y}}^{(i)}\}_{i=1}^{N_*} \right)$ is not the sum of these errors. For all the considered problems, the error is dominated by the approximation with optimal transport. 

\begin{table*}[h]
\setlength{\tabcolsep}{3pt}
\caption{RRMSE for the successive stages of TOS-GP: errors between test and approximated signals (Approximation), errors between test and predicted transferred signals (Transferred prediction), and errors between test and predicted signals (Total).}
\label{tab:detail_errors}
\centering
\begin{tabular}{ccccccc} 
 \toprule
 Stage\textbackslash Dataset & \texttt{Rotor37}(\texttt{P}) & \texttt{Rotor37}(\texttt{T}) & \texttt{Tensile2d}(\texttt{U}) & \texttt{Tensile2d}(\texttt{$\sigma_{12}$})  \\
 %& x10\textsuperscript{-3} & x10\textsuperscript{-3} & x1 & x1\\
 \midrule
 Approximation & 3.29e-2 & 9.51e-3 & 1.90e-3 & 4.55e-3 \\
 Transferred Prediction & 2.59e-2 & 2.08e-3 & 1.35e-3 & 3.37e-3 \\
 Total & 3.36e-2 & 9.63e-3 & 2.23e-3 & 5.57e-3 \\
 
 \bottomrule
\end{tabular}
\end{table*}

Table \ref{tab:H_varying} shows RRMSE scores as a function of the number of continuous WL iterations selected. This parameter has little influence on the error.

\begin{table*}[h]
\setlength{\tabcolsep}{3pt}
\caption{RRMSE scores depending on the number of continuous WL iterations.}
\label{tab:H_varying}
\centering
\begin{tabular}{ccccccc} 
 \toprule
 WL iterations\textbackslash Dataset & \texttt{Rotor37}(\texttt{P}) & \texttt{Rotor37}(\texttt{T}) & \texttt{Tensile2d}(\texttt{U}) & \texttt{Tensile2d}(\texttt{$\sigma_{12}$})  \\
 %& x10\textsuperscript{-3} & x10\textsuperscript{-3} & x1 & x1\\
 \midrule
 0 & 4.38e-2 & 9.63e-3 & 2.91e-3 & 9.60e-3 \\
 1 & 3.36e-2 & 9.82e-3 & 2.23e-3 & 6.41e-3 \\
 2 & 3.52e-2 & 1.01e-2 & 2.35e-3 & 5.57e-3 \\
 3 & 3.71e-2 & 1.04e-2 & 2.34e-3 & 5.59e-3 \\
 \bottomrule
\end{tabular}
\end{table*}

\subsection{Multiscale hyperelasticity problem: comparison with MGN}

\begin{figure*}[h!]
\centering
\includegraphics[width=0.99\textwidth]{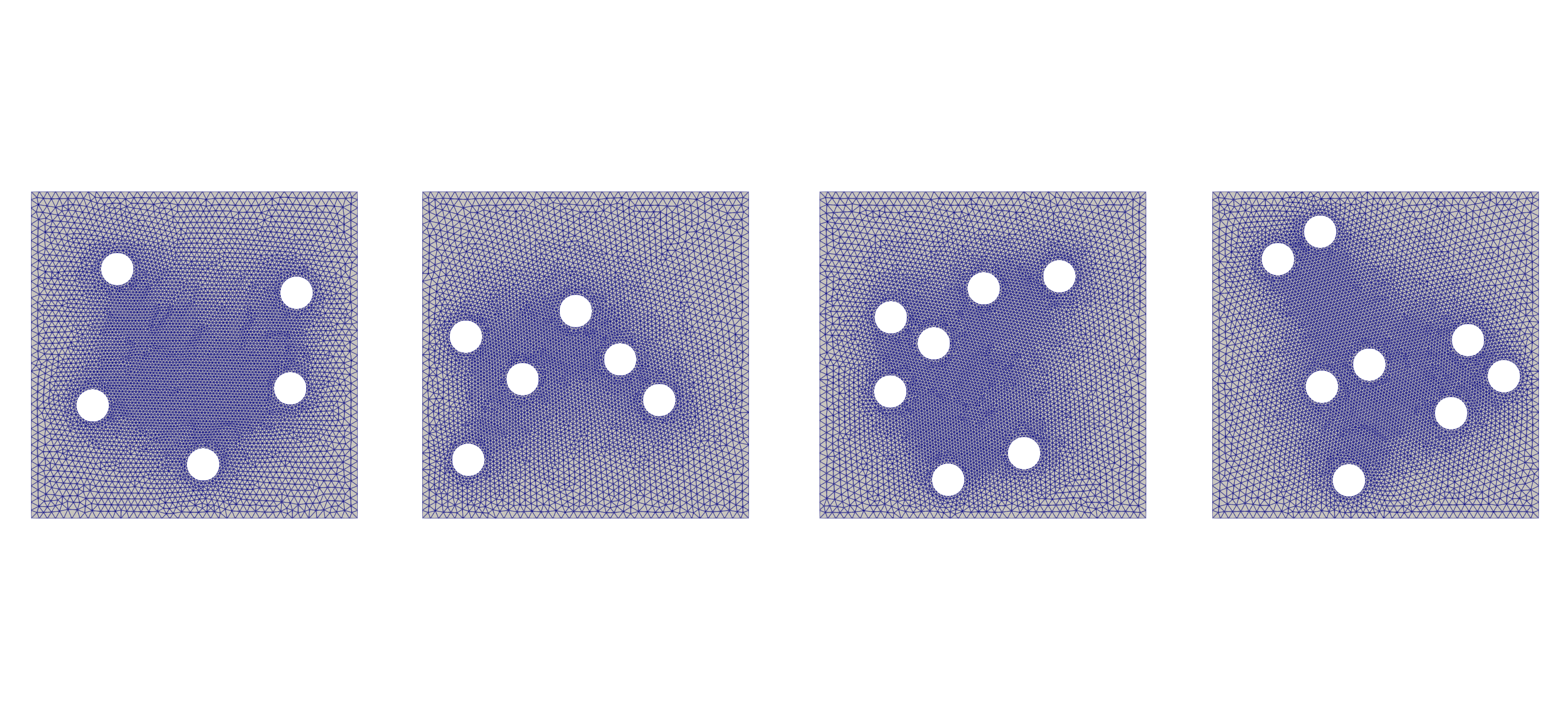}
\caption{Examples of input meshes from the multiscale problem.}
\label{fig:examples_rves}
\end{figure*}

Figure \ref{fig:examples_rves} shows four meshes corresponding to the four different topologies of the multiscale problem with $5$ to $8$ pores. Both the positions of the pores and their number can vary between samples.

\begin{figure*}[h!]
\centering
    
\includegraphics[width=0.32\textwidth]{emmentalisticity_new_0_sample12_true.png}
\includegraphics[width=0.32\textwidth]{emmentalisticity_new_0_sample12_pred.png}
\includegraphics[width=0.32\textwidth]{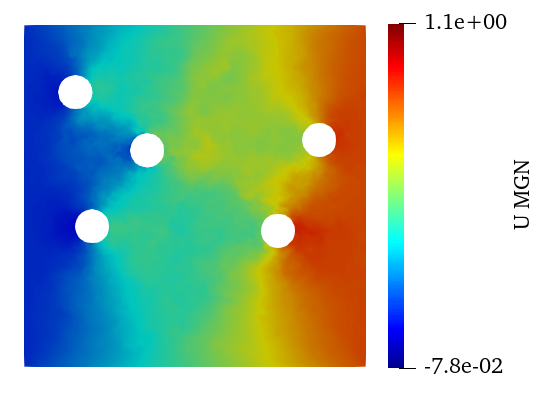}

\includegraphics[width=0.32\textwidth]{emmentalisticity_new_0_sample302_true.png}
\includegraphics[width=0.32\textwidth]{emmentalisticity_new_0_sample302_pred.png}
\includegraphics[width=0.32\textwidth]{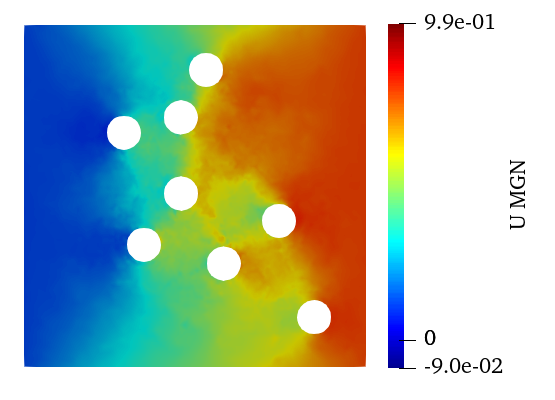}

    \caption{Two samples with different topologies from the multiscale dataset. From left to right: true field, predicted field with TOS-GP, predicted field with MGN.}
    \label{fig:emmentalisticity_MGN_vs_TOS_GP}
\end{figure*}

In addition to the results for field prediction with TOS-GP, we carried out experiments with MGN in order to compare them. We highlight the predicted fields with both methods for two samples with 5 and 7 pores respectively in Figure \ref{fig:emmentalisticity_MGN_vs_TOS_GP}. We note that predictions with MGN do not have the expected regularity of a solution to such a computational mechanics problem. Moreover, MGN predicts poorly parts of the field with high absolute values, with a kind of compression towards the mean values. Such shortcomings are absent when using TOS-GP. The previous observations are also confirmed by the global errors over the entire dataset, as the test RRMSE of MGN is equal equal to 0.053 compared to 0.044 for TOS-GP.

\end{document}